\documentclass[lettersize,journal]{IEEEtran}
\usepackage{amsmath,amsfonts}
\usepackage{algorithmicx}
\usepackage{algorithm}
\usepackage{array}
\usepackage[caption=false,font=normalsize,labelfont=sf,textfont=sf]{subfig}
\usepackage{textcomp}
\usepackage{stfloats}
\usepackage{url}
\usepackage{verbatim}
\usepackage{graphicx}
\usepackage{cite}
\usepackage{algpseudocode}
\usepackage[algo2e]{algorithm2e}
\usepackage{adjustbox}
\usepackage{booktabs}
\usepackage{xcolor}
\usepackage{multirow}
\usepackage[colorlinks=true, linkcolor=blue, anchorcolor=blue, citecolor=blue, urlcolor=blue]{hyperref}
\begin{document}

\title{BIMII-Net: Brain-Inspired Multi-Iterative Interactive Network for RGB-T Road Scene Semantic Segmentation}


\author{Hanshuo Qiu, Jie Jiang, Ruoli Yang\IEEEauthorrefmark{1}, Lixin Zhan, and Jizhao Liu%
	\thanks{Hanshuo Qiu and Jizhao Liu are with the School of Information Science and Engineering, Lanzhou University, No.222, TianShui Road(south), Lanzhou, 730000, Gansu, China (e-mail: qiuhsh21@lzu.edu.cn; liujz@lzu.edu.cn).}%
	\thanks{Jie Jiang, Ruoli Yang and Lixin Zhan are with the College of Systems Engineering, National University of Defense Technology, Changsha 410073, China (e-mail: jiejiang@nudt.edu.cn; yrl\_14@nudt.edu.cn; zhanlixin98@outlook.com).}%
	\thanks{Hanshuo Qiu and Jie Jiang are co-first authors.}%
	\thanks{\IEEEauthorrefmark{1}Ruoli Yang is the corresponding author.}
}

\markboth{Journal of \LaTeX\ Class Files,~Vol.~14, No.~8, August~2021}%
{Shell \MakeLowercase{\textit{et al.}}: A Sample Article Using IEEEtran.cls for IEEE Journals}


\maketitle

	\begin{abstract}
	
	RGB-T road scene semantic segmentation enhances visual scene understanding in complex environments characterized by inadequate illumination or occlusion by fusing information from RGB and thermal images. Nevertheless, existing RGB-T semantic segmentation models typically depend on simple addition or concatenation strategies or ignore the differences between information at different levels. To address these issues, we proposed a novel RGB-T road scene semantic segmentation network called Brain-Inspired Multi-Iteration Interaction Network (BIMII-Net). First, to meet the requirements of accurate texture and local information extraction in road scenarios like autonomous driving, we proposed a deep continuous-coupled neural network (DCCNN) architecture based on a brain-inspired model. This architecture efficiently integrates multi-scale information via iterative optimization of multiple continuous-coupled neural network (CCNN) layers and employs a phased training technique to reduce computational resource consumption while maintaining performance. Second, to enhance the interaction and expression capabilities among multi-modal information, we designed a cross explicit attention-enhanced fusion module (CEAEF-Module) in the feature fusion stage of BIMII-Net to effectively integrate features at different levels. Finally, we constructed a complementary interactive multi-layer decoder structure, incorporating the shallow-level feature iteration module (SFI-Module), the deep-level feature iteration module (DFI-Module), and the multi-feature enhancement module (MFE-Module) to collaboratively extract texture details and global skeleton information, with multi-module joint supervision further optimizing the segmentation results. Experimental results demonstrate that BIMII-Net achieves state-of-the-art (SOTA) performance in the brain-inspired computing domain and outperforms most existing RGB-T semantic segmentation methods. It also exhibits strong generalization capabilities on multiple RGB-T datasets, proving the effectiveness of brain-inspired computer models in multi-modal image segmentation tasks.
\end{abstract}


\section{Introduction}

Semantic segmentation is a fundamental task in computer vision that aims to classify each pixel in an image, thereby facilitating a profound comprehension of the image or scene at the pixel level. This task has broad implications across various domains, including autonomous driving, medical imaging, robotics, and remote sensing\cite{LI202115, 9913352, Chen_2022_CVPR, QURESHI2023316, 9561398, SU2021106418}. Traditional semantic segmentation techniques predominantly rely on convolutional neural networks (CNNs) \cite{alzubaidi2021review, 222795}. In complex scenes, RGB images frequently fail to deliver sufficient information due to issues such as variations in lighting, weather conditions, and occlusion, leading to a decline in segmentation efficacy\cite{lin2023variationalprobabilisticfusionnetwork, 10138593}. In contrast to RGB images, thermal images effectively capture the thermal radiation emitted by objects, thereby preserving robust perceptual abilities in challenging environments\cite{10483657, 9342179}. Consequently, the incorporation of thermal imagery effectively addresses this limitation, rendering RGB-T semantic segmentation an increasingly prominent research avenue.

\begin{figure*}[htbp]
	\centerline{\includegraphics[width=45pc]{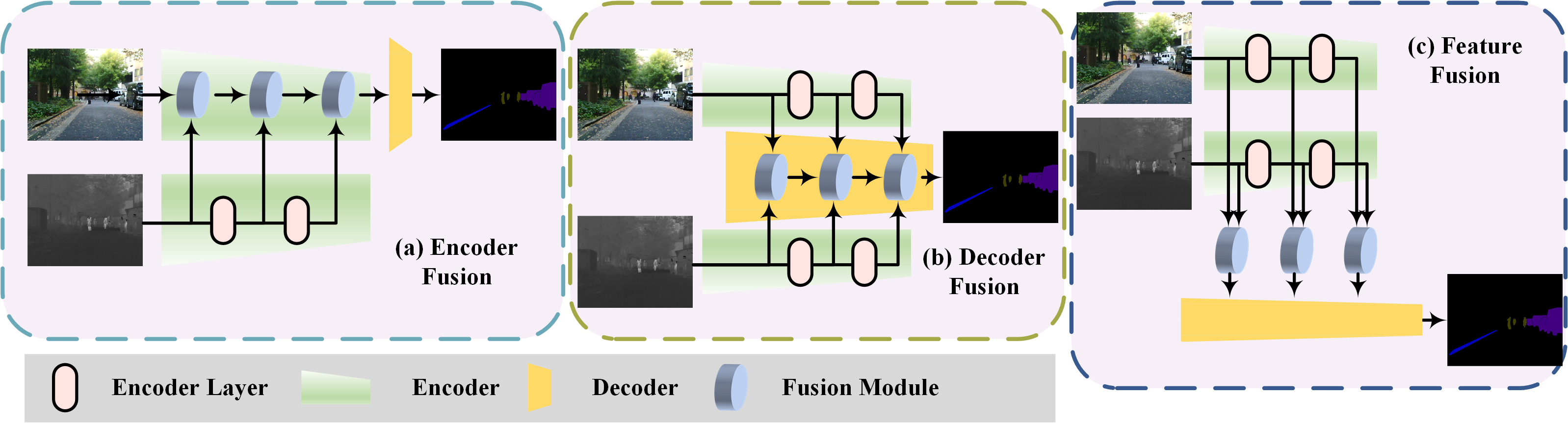}}
	\caption{Three fusion paradigms for RGB-T semantic segmentation: (a) encoder fusion, (b) decoder fusion, (c) feature fusion.
	}
	\label{fuse_en}
\end{figure*}

In recent years, RGB-T semantic segmentation methods have mainly focused on the fusion of multi-modal information, which can be roughly divided into three fusion paradigms: encoder fusion \cite{8666745, deng2021feanet, 9108585, 9447924}, decoder fusion \cite{8206396, 9578077, 9749834}, and feature fusion \cite{liu2022gcnet, 9531449}. Specifically, encoder fusion integrates multi-modal information in the feature extraction stage and pays more attention to obtaining richer details in the early stage of semantic segmentation. Decoder fusion emphasizes the integration of semantic information from different modalities throughout the upsampling and feature recovery stages. Feature fusion strategies jointly process multi-modal information in the intermediate phase. Fig. \ref{fuse_en} illustrates these three fusion paradigms. Li et al. identified the shortcomings of the above three fusion methods \cite{9900351}. Most of these fusion paradigms employ simple approaches like addition or concatenation or fail to consider the differences in information across different levels, hence constraining the further enhancement of model performance. In this study, we employ feature fusion methods and segment the decoder into three branches to address the limitations of existing fusion paradigms. Two branches consist of the shallow-level feature iteration module (SFI-Module) and the deep-level feature iteration module (DFI-Module) for processing shallow and deep features, respectively. The other branch includes the multi-feature enhancement module (MFE-Module) for integrating shallow-level and deep-level features to extract semantic segmentation results. This strategy fully considers the complementarity of information at different levels and promotes their continuous refinement.

We also consider the key role of texture and local information in feature extraction. In road scenarios with high requirements for environmental perception, such as autonomous driving, details like pedestrian stance and vehicle edges significantly impact the accuracy of segmentation results. Therefore, we desire to implement a module incorporating a feedback mechanism to perpetually refine and rectify feature expression through multiple iterations. Moreover, in complex visual tasks like semantic segmentation, the human brain excels in visual perception and adeptly processes complicated spatial and temporal relationships. In recent years, several academics have proposed various methodologies to integrate the benefits of human cognitive processes in visual perception and spatial relationship analysis into deep learning models. We introduce a novel brain-inspired model: continuous-coupled neural network (CCNN) \cite{9770397}. Although the majority of brain-inspired models prioritize minimizing power consumption and computational complexity, CCNN maintains the biological plausibility of brain-inspired computing while simultaneously delivering exceptional performance in intricate computer vision applications. We apply the multiple iteration mechanism of CCNN in the learning process of each layer in the network feature extraction and upsampling process, enabling CCNN to efficiently integrate multi-scale information. In addition, we consider that multiple iterations may introduce significant computational complexity. In order to minimize computing resource consumption while maintaining model performance, we adopted a phased training strategy. During the direct training phase, we accelerate the training by setting a lower iteration step $(T=1)$; subsequently, in the fine-tuning phase, we appropriately increase the iteration step $(T=4)$ to further optimize the expression of fine-grained features. By continuously optimizing its internal dynamics, CCNN simulates the brain's ability to perform multi-stage information processing and decision-making, thereby improving its overall performance.

In this paper, we propose a novel RGB-T road scene semantic segmentation network: Brain-Inspired Multi-Iterative Interactive Network (BIMII-Net). The overall architecture consists of three parts: encoder, feature fusion, and decoder: (a) The encoder utilizes the Segformer-B3 architecture. We employ CCNN layers in the encoder and utilize its iterative mechanism to extract features from RGB and thermal images through two branches. (b) During the feature fusion phase, we consider the three cases proposed by Liang et al \cite{10113725}. To ensure that at least one branch of feature fusion stage can obtain useful information, we use the improved cross explicit attention-enhanced fusion module (CEAEF-Module) to integrate multi-modal information. We categorize the features in the feature fusion phase into shallow-level features ($E_1$, $E_2$) and deep-level features ($E_3$, $E_4$). The shallow-level features emphasize texture and contour details, whereas the deep-level features concentrate on depicting global skeletal positioning information. (c) We designed a SFI-Module, a DFI-Module, and a MFE-Module for the decoder. SFI-Module and DFI-Module employ the iterative technique of CCNN, similar to that in the encoder, to extract texture and skeletal information. Moreover, they incorporate the final result via multiple learnable parameters of MFE-Module. Consequently, we adopt a multi-module joint supervision strategy to optimize the decoder output from three perspectives: semantic supervision, boundary supervision, and binary supervision.

The main contributions of this paper are as follows:

\begin{enumerate}
	
	\item In order to fully exploit the complementarity between features at different levels and achieve accurate extraction of fine-grained features, we propose a novel RGB-T road scene semantic segmentation network, BIMII-Net, based on a brain-inspired computing architecture. To the best of our knowledge, it is the first time that a brain-inspired computing architecture has been used to implement semantic segmentation of multi-modal data. Our model demonstrates superior performance compared to the state-of-the-art (SOTA) architectures in the domain of brain-inspired computing.
	
	\item We proposed a deep continuous-coupled neural network (DCCNN) architecture with cross-layer signal transmission characteristics. The architecture consists of multiple CCNN layers, forming an overall structure that is iteratively optimized layer by layer and continuously integrates input features from different scales during the recursive update process. In addition, the multi-stage training strategy effectively reduces the resource consumption caused by multiple iterations, thereby facilitating efficient fusion of deep features while maintaining high performance.
	
	\item We proposed a CEAEF-Module, which enhances the interaction among various modalities and the expressiveness of output features. Furthermore, the module employs two complementary branches to guarantee that at least one of them can be completely activated.
	
	\item We proposed a complementary interactive multi-layer decoder structure. The decoder structure consists of three main modules: SFI-Module is used to extract texture details; DFI-Module focuses on global skeleton information; MFE-Module realizes the complementary interaction of multi-level features.
	
\end{enumerate}

The rest of this paper is organized as follows: In Sec. \ref{secc}, we present the related work. Sec. \ref{sec3} discusses the overall architecture of BIMII-Net and introduces the details of each key module. Sec. \ref{sec4} covers the experimental results and analysis. Finally, Sec. \ref{sec5} concludes this paper.

\section{Related work}\label{secc}

\subsection{RGB Semantic Segmentation}

The rapid advancement of semantic segmentation has led to the proposal of numerous traditional network architectures. The initial Fully Convolutional Network (FCN) established the groundwork for deep learning in semantic segmentation by using transposed convolution and skip connections\cite{Long_2015_CVPR}. Subsequently, U-Net excelled in domains like medical image segmentation because of its symmetrical encoding-decoding structure and effective feature fusion mechanism\cite{10.1007/978-3-319-24574-4_28}. Moreover, the DeepLab series of networks improved the modeling of contextual information using dilated convolution and Atrous Spatial Pyramid Pooling (ASPP), while the implementation of conditional random fields further refined the segmentation boundaries\cite{7913730}. SegNet is designed to be lightweight and efficient, using pooling indexes to preserve spatial detail, rendering it an optimal selection for resource-constrained environments\cite{7803544}.

Transformer-based models have demonstrated significant potential in semantic segmentation tasks in recent years\cite{XIAO2023104791}. The Vision Transformer (ViT) was the first to incorporate the self-attention mechanism into the segmentation work for the purpose of modeling global context information\cite{10.1145/3505244}. The Swin Transformer dramatically decreased computational costs by hierarchical windowed self-attention\cite{Liu_2021_ICCV}. Moreover, Segmenter enhances the precision of segmentation boundaries by the integration of transformer and decoder modules\cite{Strudel_2021_ICCV}. SegFormer enhanced segmentation performance by a lightweight design and efficient decoders\cite{NEURIPS2021_64f1f27b}.

RGB-based semantic segmentation algorithms encounter difficulties in complex situations, including susceptibility to lighting fluctuations and challenges in differentiating items with analogous colors. To address these issues, RGB-T semantic segmentation utilizes the complementary information gathered from thermal images, which are resilient to varying lighting conditions and capture distinctive thermal radiation characteristics. This integration significantly enhances segmentation performance in challenging environments and expands the applicability of semantic segmentation.

\subsection{RGB-T Semantic Segmentation}

In the field of RGB-T semantic segmentation, a variety of models have emerged in recent years. Ha et al. first proposed a network MFNet that combines visible light and thermal images for semantic segmentation. The network adopts an encoder-decoder architecture and improves the semantic segmentation performance in complex scenes through multi-scale feature extraction and modal feature fusion\cite{8206396}. Sun et al. used two ResNet branches as encoders and proposed RTFNet\cite{8666745}. The encoder part of the network uses additive interaction to effectively combine the features of RGB and thermal images to improve segmentation performance. Sun et al. adopted an architecture similar to RTFNet and proposed the FuseSeg model by replacing ResNet with DenseNet\cite{9108585}. Shivakumar et al. proposed the PST900 model, which achieved multi-modal semantic segmentation with higher inference speed\cite{LIANG20239}.

However, the fusion strategies employed by these networks are relatively simple, mostly depending on the addition or concatenation of features, consequently hindering a comprehensive exploration of the complementarity inherent in multi-modal information. To address these problems, some models have begun to introduce more complex feature fusion mechanisms in recent years and jointly supervise features of different forms. Deng et al. proposed FEANet, which uses the feature-enhanced attention module to enhance the complementarity of multi-level features and modalities from the perspective of channels and space\cite{deng2021feanet}. Zhou et al. proposed MFFENet, which achieved multi-modal fusion at multiple stages\cite{9447924}. This model introduces the spatial attention mechanism module to make the network pay more attention to the foreground object. Zhang et al. proposed ABMDRNet, which uses a modality difference reduction and fusion subnetwork to reduce modality difference\cite{9578077}. Zhou et al. proposed GMNet, which uses a shallow feature fusion module and a deep feature fusion module to fuse features at different levels and supervises the network using three loss functions\cite{9531449}. Liu et al. proposed GCNet, which adopts a grid-like context-aware module to capture rich semantic information from different contexts\cite{liu2022gcnet}. Wang et al. proposed UTFNet, which designed an uncertainty estimation and evidence fusion module to quantify the uncertainty of each modality and used uncertainty to guide information fusion\cite{10273407}. Zhou et al. proposed CACFNet, which introduced a cross-modal attention fusion module to combine complementary information from two modalities and designed a region-based module to explore the relationship between regions and pixels\cite{10251592}. Zhou et al. proposed MMSMCNet, which uses a multi-modal complementary supervision mechanism to optimize target objects of different forms at multiple scales\cite{10123009}. Liang et al. proposed EAEFNet and designed explicit attention-enhanced fusion\cite{10113725}. Lv et al. proposed CAINet\cite{10379106}. The model uses a context-aware complementary reasoning module to establish a complementary relationship between multi-modal features and long-term context in spatial and channel dimensions.

Among the existing RGB-T semantic segmentation architectures, few methods pay attention to the differences between different morphological features or adopt advanced fusion strategies to fully consider the situation when none of the fusion branches can obtain effective information. In addition, existing methods generally lack a top-down feedback regulation mechanism similar to the human brain's visual mechanism, hindering the continuous optimization of boundaries and local details. In this study, we proposed a complementary interactive multi-layer decoder structure to enhance the complement of boundary and skeleton information. Moreover, we proposed a cross explicit attention-enhanced fusion module (CEAEF-Module) to fuse the features of each layer in the encoder. The module constructs two complementary activation branches to guarantee that at least one branch can obtain useful information. Furthermore, we introduced a continuous-coupled neural network (CCNN) with high biological plausibility and employed its iterative feedback mechanism within the encoder and decoder to continuously refine and optimize multi-scale features. To further reduce the complexity of model debugging, we adopted an automatic weighted loss (AWL) strategy. We consider several supervision tasks as components of multi-task learning and use learnable parameters to dynamically modify the weights of each loss, therefore achieving a more reasonable weight distribution and enhancing the model's performance \cite{Kendall_2018_CVPR}.

\subsection{Brain-Inspired Computing in Semantic Segmentation}

In recent years, brain-inspired computing has emerged as a significant area of focus within computational neuroscience and artificial intelligence\cite{10636118, Liu2024AdvancingBC}. Nonetheless, initial applications of brain-inspired computing primarily concentrated on straightforward tasks like image classification, which typically do not necessitate intricate spatial processing of input data\cite{10378944, robertson2020ultrafast}. In recent years, the advancement of brain-inspired computing theory and technology has led to an expansion of research from image classification to more complex tasks, including semantic segmentation and object detection\cite{9306772}. Semantic segmentation requires that the network possesses both global semantic understanding and local detail capture capabilities. This presents new challenges for the application of brain-inspired computing and fostering a range of innovative approaches.

Recently, researchers have proposed a variety of brain-inspired computing architectures to meet these challenges. Kim et al. first applied spiking neural network (SNN) to semantic segmentation tasks\cite{kim2022beyond}. The study implemented a low-power semantic segmentation model and used a proxy gradient optimization method to obtain better performance and lower latency. Yao et al. proposed a spike-driven Transformer and designed a Meta Transformer-based SNN architecture\cite{yao2024spikedriven}. This architecture is the first directly trained SNN architecture that can handle classification, detection, and segmentation simultaneously, and this study is the first time that the performance of the SNN field has been reported on the ADE20K dataset. Zhou et al. proposed QKFormer and designed a new SNN formal attention mechanism\cite{zhou2024qkformer}. This is a QK attention mechanism in the form of a spike, which has linear complexity and is suitable for the spatiotemporal spike pattern of SNN. Li et al. proposed Spike Calibration and used the artificial neural network (ANN) to SNN method for training\cite{li2022spikecalibrationfastaccurate}. This study used spike calibration (SpiCalib) to eliminate the damage of discrete pulses to the output distribution, and only required extremely short inference time and extremely low energy consumption to obtain effective segmentation and detection effects.

Currently, semantic segmentation models based on brain-inspired computing still face many limitations. To the best of our knowledge, there has been no proposed effective brain-inspired computing architecture for the semantic segmentation of multi-modal data. Current brain-inspired computer architectures for semantic segmentation are often compared to SNNs and underperforming ANNs, indicating that they have yet to attain the standards of prevalent deep learning methods. Moreover, many brain-inspired models exhibit a deficiency in biological rationality, and their biological foundations are inadequately aligned with actual issues, thereby constraining the efficacy and advancement of brain-inspired computer models in practical applications.

\subsection{Continuous-Coupled Neural Network}\label{CCNNsec}

CCNN is an innovative artificial neural network model inspired by the primary visual cortex of mammals, such as cats\cite{9770397}. The model is designed to more precisely simulate the dynamic response of actual neurons to intricate external inputs, encompassing both periodic and chaotic behaviors. In comparison to the conventional pulse-coupled neural network (PCNN), the CCNN more accurately reflects the electrophysiological properties of biological neurons and has superior efficacy in image and video processing tasks\cite{lindblad2005image, electronics11203264}. Currently, CCNN is widely used in various other domains like brain-inspired computing and dynamic visual processing, offering a new perspective for investigating how the brain employs chaos to process information\cite{mi14112113, Zhang_2024, zhang2024revealing, wang2024chaosmotionunveilingrobustness}.

The structure of CCNN is shown in Fig. \ref{DCCNN}, and its mathematical iteration equation is as follows:

\begin{equation}
	\begin{cases} 
		F_{ij}(n) = e^{-\alpha_f} F_{ij}(n-1) + V_F M_{ijkl} Y_{kl}(n-1) + S_{ij} \\ 
		L_{ij}(n) = e^{-\alpha_l} L_{ij}(n-1) + V_L W_{ijkl} Y_{kl}(n-1) \\ 
		U_{ij}(n) = F_{ij}(n) \big( 1 + \beta L_{ij}(n) \big) \\ 
		Y_{ij}(n) = \frac{1}{1 + e^{-\big(U_{ij}(n) - E_{ij}(n)\big)}} \\ 
		E_{ij}(n) = e^{-\alpha_e} E_{ij}(n-1) + V_E Y_{ij}(n-1)
	\end{cases}
	\label{CCNNe}
\end{equation}

where $(i, j)$ represents the position of the neuron within the image matrix. CCNN contains five main parts: coupled connection $L_{ij}(n)$, feedback input $F_{ij}(n)$, modulation product $U_{ij}(n)$, dynamic threshold $E_{ij}(n)$, and continuous output $Y_{ij}(n)$. $S_{ij}$ denotes the external input received by the receptive field. Parameters $a_f$, $a_l$, and $a_e$ represent the exponential decay factors of the neuron. $V_F$ and $V_L$ represent the weighting factors of the neuron action potential, respectively. In addition, $M_{ijkl}$ and $W_{ijkl}$ denote the feedback and connection synaptic weights, respectively, and $\beta$ represents the connection strength, which directly influences $L_{ij}(n)$ in the modulation product $U_{ij}(n)$.


In the field of brain-inspired computing, SNN is regarded as the third generation of neural network models, replacing traditional continuous signals with discrete pulses \cite{MAASS19971659}. It has been recognized for lower power consumption and computational complexity compared to traditional ANN models, making it particularly suitable for embedded systems and real-time applications \cite{10181703, 10447121}. Nonetheless, its discrete pulse outputs frequently result in issues such as non-differentiability and information loss. In contrast, CCNNs utilize continuous outputs, which partially address these deficiencies and demonstrate enhanced biological plausibility and performance in visual tasks. However, due to the continuous signal outputs combined with convolution operations, their overall energy consumption is comparatively elevated, potentially posing challenges in large-scale application scenarios.

The initial implementation of CCNN in deep learning has been reported in Ref. \cite{zhang2024revealing}. Zhang et al. employed the deep continuous-coupled neural network (DCCNN) for image classification and identified the phenomenon of semantic saturation in CCNN. This study employs CCNN for the semantic segmentation of multi-modal data. This is the first application of CCNN for the intricate computer vision task of multi-modal semantic segmentation.

\section{Proposed Method}\label{sec3}

\subsection{Overall Architecture}

\begin{figure*}[htbp]
	
	\centerline{\includegraphics[width=45pc]{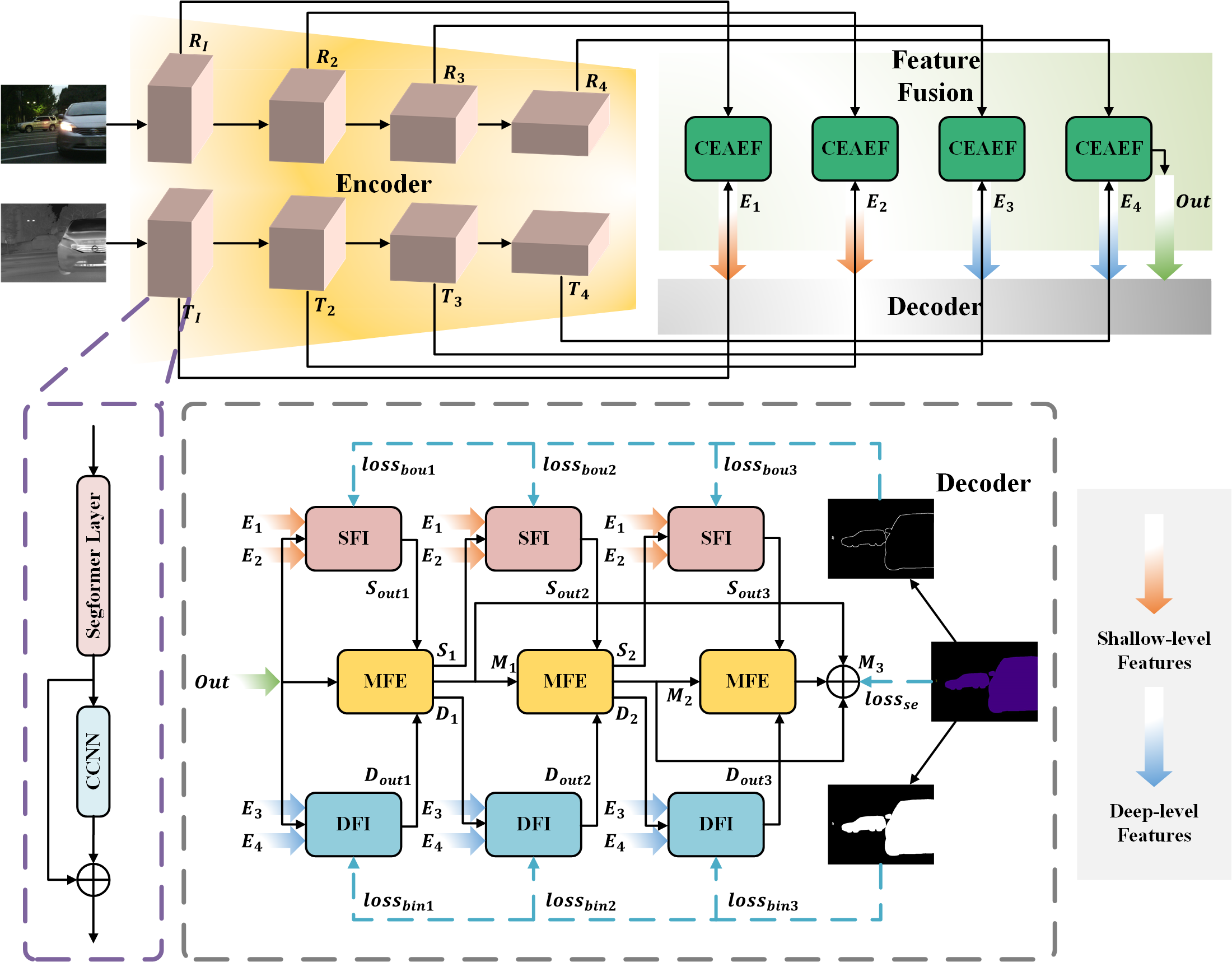}}
	\caption{Overall architecture of proposed BIMII-Net. BIMII-Net consists of an encoder, a feature fusion part, and a decoder. The encoder is based on the Segformer structure. A CCNN layer is added after each Segformer layer and fused with the output through a residual connection to further iterate features. This part extracts multi-scale features from RGB and thermal images. The feature fusion part fuses the features from the encoder through the CEAEF-Module. The initial inputs in the decoder are generated in the section. The decoder further separates shallow-level features and deep-level features, extracting texture and skeleton information through the SFI-Module and DFI-Module, respectively. Subsequently, the MFE-Module fuses them to generate the final semantic segmentation result. Furthermore, the overall architecture is optimized through a multi-module joint supervision strategy.
	}
	\label{BIMII-Net}
\end{figure*}

As shown in the Fig. \ref{BIMII-Net}, BIMII-Net consists of three parts: encoder, feature fusion, and decoder.

Encoder: We adopt the encoder structure based on Segformer\cite{xie2021segformer}. Following each Segformer layer, a continuous-coupled neural network (CCNN) layer is added, fused through a residual connection. We input RGB and thermal images into two branches for feature extraction, obtaining the outputs $R_i$ and $T_i$ from the i-th $(i \in \{1, 2, 3, 4\})$ CCNN layer in both branches of the encoder. The outputs are sent to the feature fusion part.

Feature fusion: We utilize the cross explicit attention-enhanced fusion module (CEAEF-Module) to perform feature fusion on $R_i$ and $T_i$ and produce $E_i$ as output. The outputs of the four CEAEF-Modules are categorized into two categories: shallow-level features ($E_1$, $E_2$) and deep-level features ($E_3$, $E_4$). The initial input for the decoder utilizes the output from the last CEAEF-Module.

Decoder: We design a complementary interactive multi-layer decoder structure. The shallow-level feature iteration module (SFI-Module) and the deep-level feature iteration module (DFI-Module) in the decoder are used to attain the localization of the target contour and skeleton, respectively. Their outputs are denoted as $S_{outj}$ and $D_{outj}$ $(j \in \{1, 2, 3\})$, respectively. The multi-feature enhancement module (MFE-Module) integrates the fused features to provide the final semantic segmentation result, comprising three distinct outputs, denoted as $S_j, M_j$, and $D_j$. The three outputs are used as the inputs for the subsequent SFI-Module, MFE-Module, and DFI-Module, respectively. Eventually, we implement a multi-module joint supervision strategy to jointly supervise the outputs of all modules within the decoder. $S_{outj}$ and $D_{outj}$ are supervised for boundary and binary, respectively, and $M_{out}$ is supervised for semantic segmentation.

\subsection{Deep Continuous-Coupled Neural Network}

\begin{figure}[htbp]
	
	\centerline{\includegraphics[width=22pc]{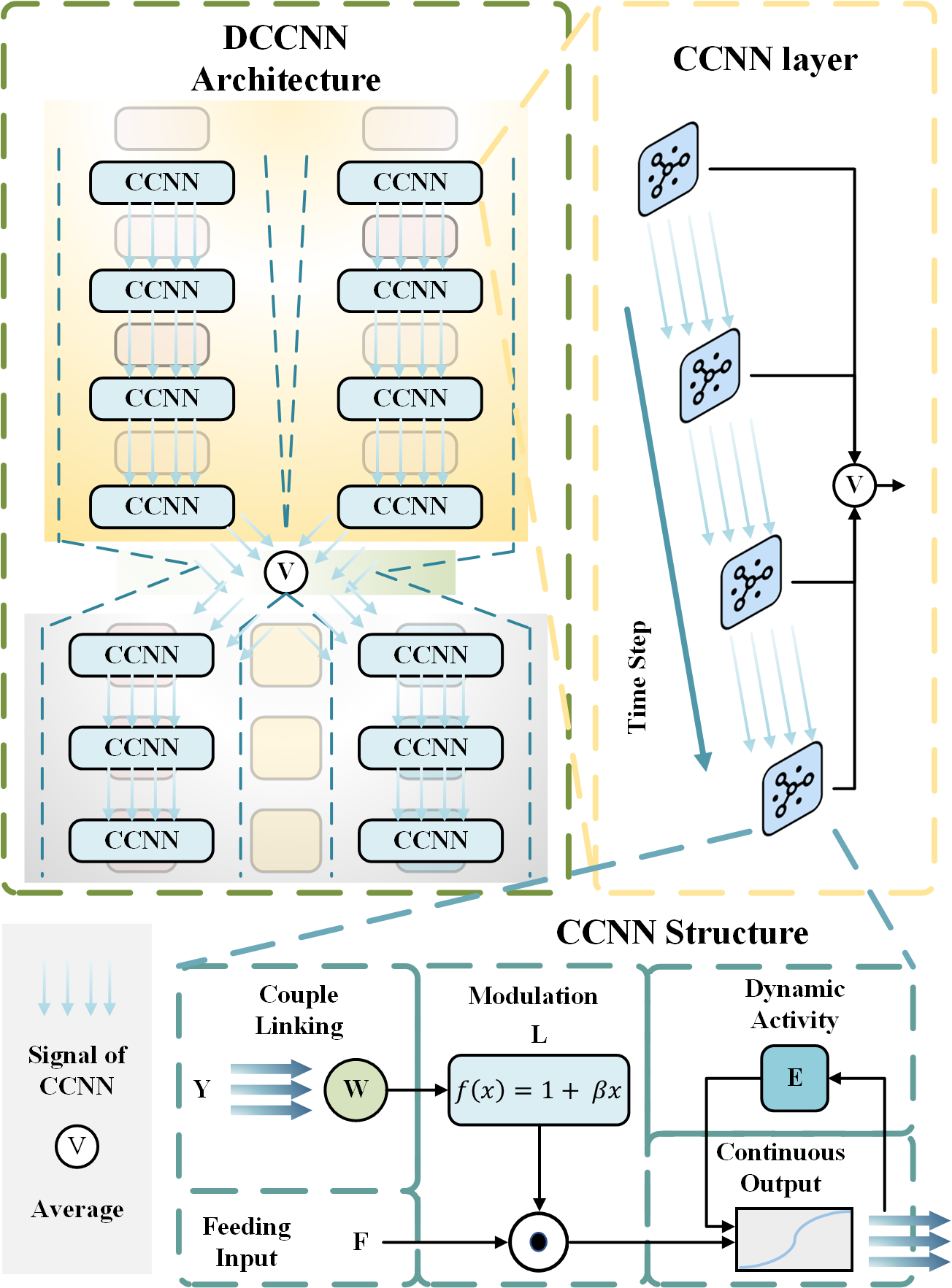}}
	\caption{The overview of the proposed DCCNN architecture. The DCCNN architecture consists of 7 CCNN layers, with the encoder comprising 4 CCNN layers and the decoder including 3 CCNN layers. Signal transmission between layers is facilitated through state parameters. Each CCNN layer processes information across multiple time steps, and its output is obtained by averaging the results of these time steps. Additionally, during the feature fusion stage, the signals from the final encoder layer are averaged before being passed on to the decoder.
	}
	\label{DCCNN}
\end{figure}

In the proposed BIMII-Net, CCNN is embedded in the model as a deep continuous-coupled neural network (DCCNN) architecture, which is shown in Fig. \ref{DCCNN}. In the proposed architecture, the encoder contains four CCNN layers, while the decoder includes three CCNN layers. Each CCNN layer updates iteratively using the state parameters from the previous layer. Furthermore, we compute the average of the signals (such as $F, L, E$) from the last CCNN layer in both branches of the encoder in the feature fusion stage and transmit them to the decoder as the initial input for the next CCNN layer. When $T = 4$, the CCNN in the entire network undergoes 28 iterations, processing 7 distinct feature inputs. This multi-iteration strategy not only effectively emulates the multi-stage dynamic processing of information in biological visual systems but also facilitates more refined feature optimization.

At present, most brain-inspired models need to initialize their neuron states in each layer call. The proposed scheme initializes only the first layer of the CCNN, with each subsequent layer utilizing the state parameters (such as $F, L, E$) from the preceding layer's iteration for recursive updates. In the first layer of the CCNN within our model, all neurons remain inactive, and the weights of the convolutional layer are initialized. The CCNN iterative algorithm ($T = 4$) in the RGB branch of the encoder is shown in Alg. \ref{alg}.

In deep learning, $F, L, U, Y, E$ are all Torch tensors of the same size as the external input $S$ to ensure dimensional consistency and computational efficiency. In addition, the matrices $M$ and $W$ can be substituted with convolutional layers. This design preserves the bio-inspired attributes of the network structure while markedly enhancing the model's trainability and flexibility. Moreover, the characteristic of weight sharing significantly decreases the parameter count and enhances computation efficiency, rendering CCNN more suitable for large-scale image and video processing tasks. Currently, most CCNN applications use convolutional layers instead of two matrices\cite{mi14112113, zhang2024revealing, wang2024chaosmotionunveilingrobustness}.

From a biological perspective, CCNN can simulate the chaotic dynamic behavior of biological neurons, which is fundamental to the human brain's information processing and cognitive tasks. From a deep learning perspective, the dynamic attributes of CCNN enhance nonlinear expression capabilities. Moreover, the continuously differentiable output enhances the adaptability of the brain-inspired model to the gradient optimization demands of deep learning. In the proposed model, CCNN employs a structure similar to RNN for iterative updates in both the encoder and decoder. The dynamic threshold $E_{ij}(n)$ enables CCNN to capture the temporal attributes of features and employ memory mechanisms for enhanced predictive accuracy.


\begin{algorithm}[htbp]
	\caption{CCNN iterative algorithm in the RGB branch of the encoder.}
	\label{alg}
	\KwIn{$RGB$}
	\KwOut{$Out_1, Out_2, Out_3, Out_4$}
	
	\textbf{Initialization:} $F(0) = 0$, $L(0) = 0$, $U(0) = 0$, $E(0) = 0$, $Y(0) = 0$, initialize the convolutional layers ${Conv}_M$ and ${Conv}_W$ \\
	\SetKwFunction{CCNN}{CCNN} 
	\SetKwProg{Fn}{Function}{:}{}
	\Fn{\CCNN{$X, F, L, E, Y, {Conv}_M, {Conv}_W$}}{
		$F = e^{-\alpha_f} F + {Conv}_M(Y) + X$ \\
		$L = e^{-\alpha_l} L + {Conv}_W(Y)$\\
		$U = F(n) \odot (1 + \beta L)$\\
		$E = e^{-\alpha_e} E + V_E Y$\\
		$Y = \frac{1}{1 + e^{-(U - E)}}$\\
		\Return{$F, L, E, Y, {Conv}_M, {Conv}_W$}\;
	}
	\SetKwFunction{CCNNT}{{CCNN}\_T} 
	\SetKwProg{Fn}{Function}{:}{}
	\Fn{\CCNNT{$X, F, L, E, Y, {Conv}_M, {Conv}_W$}}{
		\For{$t = 1, 2, 3, \dots, T$}{
			$F, L, E, Y, {Conv}_M, {Conv}_W =\\CCNN(X, F, L, E, Y, {Conv}_M, {Conv}_W)$ \\
			$Y^{t} = Y$
		}
		$y^{T} = \frac{1}{T} \sum_{t=1}^{T} Y^{t}$ \\
		\Return{$y^{T}, F, L, E, Y, {Conv}_M, {Conv}_W$}\;
	}
	\textbf{Main:}\\
	$X_1 = \text{{Segformer}\_{layer1}}(RGB)$\\
	$Out, F, L, E, Y, {Conv}_M, {Conv}_W =\\\CCNNT(X_1, F, L, E, Y, {Conv}_M, {Conv}_W)$\\
	$Out_1 = Out + X_1$\\
	$X_2 = \text{{Segformer}\_{layer2}}(Out_1)$\\
	$Out, F, L, E, Y, {Conv}_M, {Conv}_W =\\\CCNNT(X_2, F, L, E, Y, {Conv}_M, {Conv}_W)$\\
	$Out_2 = Out + X_2$\\
	$X_3 = \text{{Segformer}\_{layer3}}(Out_2)$\\
	$Out, F, L, E, Y, {Conv}_M, {Conv}_W =\\\CCNNT(X_3, F, L, E, Y, {Conv}_M, {Conv}_W)$\\
	$Out_3 = Out + X_3$\\
	$X_4 = \text{{Segformer}\_{layer4}}(Out_3)$\\
	$Out, F, L, E, Y, {Conv}_M, {Conv}_W =\\\CCNNT(X_4, F, L, E, Y, {Conv}_M, {Conv}_W)$\\
	$Out_4 = Out + X_4$\\
\end{algorithm}

Traditional CCNN models normally use the results of the last iteration as output. However, employing simply the results of the latest iteration as the output of the CCNN layer can only reflect the feature extraction capacity of CCNN at a single time point while ignoring the important information that the features may include at each time point. The primary visual cortex of the human brain has a similar processing method. While receiving external visual stimuli, the primary visual cortex gradually analyzes and integrates visual information through multi-stage and multi-time step activities\cite{jia2022multi, knudstrup2024learned}. Consequently, utilizing CCNN to process information across multiple temporal dimensions can more comprehensively simulate the processing mechanisms of the biological visual system, therefore improving the model's ability in understanding and utilizing input data. The output of the CCNN layer utilized is presented in Eq. \ref{iter}:

\begin{equation}
	y^{T} = \frac{1}{T} \sum_{t=n}^{n+T} Y^{t}(x)
	\label{iter}
\end{equation}

where $n$ represents the initial number of iterations when entering the CCNN layer, $T$ represents the iteration time of each CCNN layer. In the proposed model, we choose $T = 4$. $y^{T}$ denotes the final output of the CCNN layer.

\subsection{Cross Explicit Attention-Enhanced Fusion}

In multi-modal data fusion, the characteristics of different modalities often exhibit substantial disparities, making effective cross-modal feature interaction and fusion essential. Inspired by the EAEF module, we propose a novel CEAEF-Module \cite{10113725}. The primary changes are as follows:

\begin{enumerate}
	
	\item Both branches in the CEAEF-Module use the channel attention (CA). Compared with the original EAEF module, nonlinear interactions are introduced into the complementary branch to improve the data expression capability. This design enhances the alignment of subsequent features used for crossover and avoids uneven feature distribution.
	
	\item Proposed a crossover mechanism. The interaction between RGB and thermal images not only depends on the weight adjustment of CA but also introduces explicit cross-modal deep interaction within the branch. This design significantly enhances the complementarity between modalities, allowing features of different modalities to guide and optimize. The cross mechanism, in conjunction with deep convolution, enhances the representation of interactive features and fully utilizes the complementary information between multiple modalities.
	
	\item Using the CBL module to adjust the dimension at the end of the module. Different from EAEFNet, which directly inputs the fused output to the next layer of the encoder, we divide the result into two levels and output it to the decoder. This design allows detailed information, such as contours and skeletons, to be better preserved, which is more suitable for the multi-module joint supervision strategy we adopt.
	
\end{enumerate}

\begin{figure*}[htbp]
	
	\centerline{\includegraphics[width=35pc]{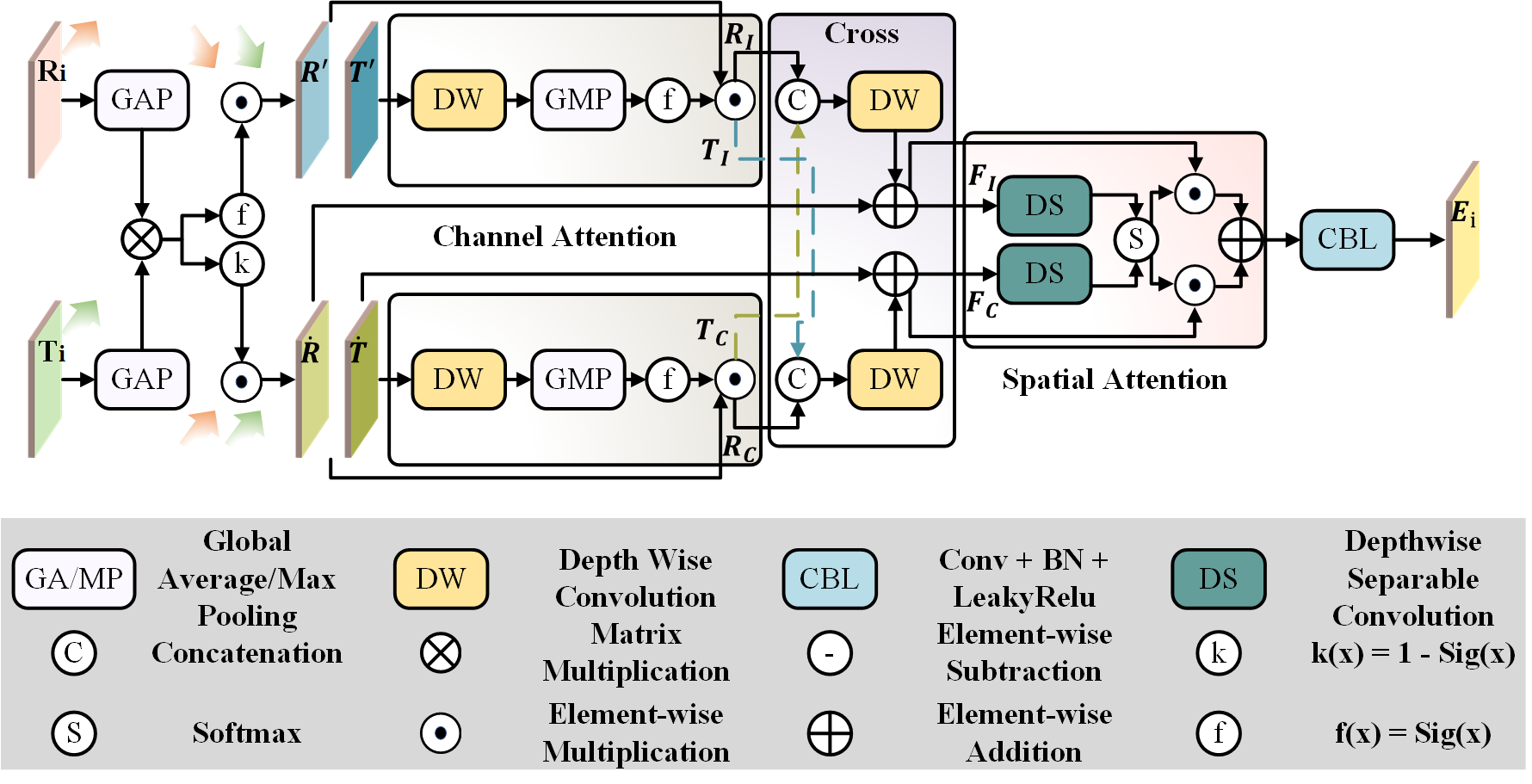}}
	\caption{Proposed CEAEF-Module. The CEAEF-Module consists of two branches, processing RGB and thermal image features, respectively. The introduction of channel attention enhances feature expression capabilities and aligns distributions. The module employs an explicit cross-modal deep interaction approach to improve feature representation and enhance complementarity across modalities. The features are ultimately integrated through a combination of deep separable convolution and spatial attention.
	}
	\label{CEAEF}
\end{figure*}

The architecture of CEAEF-Module is shown in Fig. \ref{CEAEF}. CEAEF-Module consists of two branches, and the inputs are the outputs of the CCNN layer in the encoder, denoted as $R_i$ and $T_i$. Firstly, we extract the CA of the two features as $R$ and $T$, then multiply them by the dimension $c$ and use the Sigmoid function for activation. In the first branch, the activation result $(Sig(c \cdot R \ \times T))$ is directly applied to $R_i$ and $T_i$, where $Sig$ represents Sigmoid activation. In the second branch, in order to avoid insufficient activation, $1 - Sig(c \cdot R \ \times T)$ is multiplied with $R_i$ and $T_i$ for supplementation. This design ensures that the features in the two branches are fully activated in at least one branch, which improves the integrity and robustness of feature fusion. These processes are as follows:

\begin{equation}
	\begin{aligned}
		R &= MLP(GAP(R_i))\\
		T &= MLP(GAP(T_i)) \\
		R' &= Sig(c \cdot  R \times T) \cdot R_i\\
		T' &= Sig(c \cdot R \times T) \cdot T_i \\
		\dot R &= 1 - Sig(c \cdot  R \times T) \cdot R_i \\
		\dot T &= 1 - Sig(c \cdot R \times T) \cdot T_i
	\end{aligned}
	\label{1}
\end{equation}

where GAP represents the global average pooling operation. To prevent the addition results from two branches from offsetting the Sigmoid activation, the original EAEF module employs CA in one branch. The proposed CEAEF-Module uses CA on both branches, which improves the balance and stability of the subsequent cross mechanism. The cross mechanism is set subsequent to the CA. The features of the different modals in the two branches are concatenated, followed by refinement through the depthwise convolution layer. Subsequently, the results are added to the features after the first activation in the supplementary branch. The process of the cross mechanism is as follows:

\begin{equation}
	\begin{aligned}
		\begin{bmatrix}
			Gate_{RI} \\ Gate_{TI}
		\end{bmatrix} &= Sig(MLP(GMP(DW(CAT(R', T'))))) \\
		\begin{bmatrix}
			Gate_{RC} \\ Gate_{TC}
		\end{bmatrix} &= Sig(MLP(GMP(DW(CAT(\dot{R}, \dot{T}))))) \\
		R_I &= R' \cdot Gate_{RI}\\
		\quad T_I &= T' \cdot Gate_{TI} \\
		R_C &= \dot{R} \cdot Gate_{RC}\\
		\quad T_C &= \dot{T} \cdot Gate_{TC} \\
		F_I &= DW(CAT(R_I, T_C)) + \dot{R}\\
		\quad F_C &= DW(CAT(R_C, T_I)) + \dot{T}
	\end{aligned}
	\label{2}
\end{equation}

where DW denotes depthwise convolution, CAT represents the concatenation operation, and GMP represents the global maximum pooling operation. $(R_I, T_I)$ and $(R_C, T_C)$ denote the outputs of CA, while $(F_I, F_C)$ represent the outputs of the cross mechanism. Finally, we use spatial attention (SA) to integrate the features. The traditional $1 \times 1$ convolution in spatial attention is substituted with a depthwise separable convolution, therefore increasing the capacity to extract the spatial information while diminishing the parameter number and computational complexity. The process of SA is as follows:

\begin{equation}
	\begin{aligned}
		{F_I}' &= DS(F_I)\\
		{F_C}' &= DS(F_C) \\
		\begin{bmatrix}
			{V}_{FI} \\ {V}_{FC}
		\end{bmatrix} &= Soft(CAT({F_I}', {F_C}'))) \\
		{E}_{i} &= {V}_{FI} * F_I + {V}_{FC} * F_C
	\end{aligned}
	\label{3}
\end{equation}

where DS represents depthwise separable convolution, Soft represents the Softmax operation, $({V}_{FI}, {V}_{FC})$ represents the result of the Softmax operation, and ${E}_{i}$ represents the final output of the CEAEF-Module.

\subsection{Complementary Interactive Multi-Layer Decoder}

\subsubsection{Shallow-level Feature Iteration Module}

We propose a novel SFI-Module to explore texture information in shallow-level features, and its architecture is shown in Fig. \ref{Module} (a). SFI-Module receives three inputs: $E_1, E_2$, and $S_j$ (the input of the first layer uses $E_4$ instead of $S_j$). Firstly, the three features are upsampled and adjusted to the same size by the CBL module. Since the shallow-level features are rich in detailed information and similar, we use element-by-element addition to effectively preserve the modal features and enhance the expression of shared features. For $E_1$ and the output of the previous MFE-Module $S_j$, the two features are quite different, so a concatenation operation is required to preserve the independent information of each modality. Subsequently, we employ the CBL module to extract more information and stabilize values, followed by a concatenation operation to process the combined features. These processes are as follows:

\begin{equation}
	\begin{aligned}
		F_{add} &= CBL(E_1) + CBL(E_2)\\
		F_{cat} &= CAT(CBL(E_1), CBL(S_j)) \\
		F_{fuse} &= CAT(CBL(F_{add}), CBL(F_{cat}))
	\end{aligned}
	\label{4}
\end{equation}
Subsequently, we adopted the separable convolution module with a CCNN layer. CCNN layer enhances the feature expression by multiple iterations. After the CCNN layer, we use pointwise convolution and depthwise separable convolution to enhance feature extraction efficiency and further optimize feature representation. The residual connection after the convolution layer accelerates model convergence and preserves the integrity of the original information. The process of the separable convolution module is as follows:

\begin{equation}
	\begin{aligned}
		F_{i}, L_{i}, Y_i, E_i &= CCNN(F_{i-1}, L_{i-1}, E_{i-1}, F_{fuse}) \\
		&\quad \text{for } i \in \{1, \dots, T\} \\
		{CCNN}_{out} &= \frac{1}{T} \sum_{i=1}^{T} Y_i \\
		F_{out} &= DS(PW({CCNN}_{out})) + F_{fuse}
	\end{aligned}
	\label{ccnn}
\end{equation}

where PW stands for pointwise convolution. We then use multi-scale dilated feature extraction (MDFE) to enhance texture and boundary information. Shallow-level features retain more local details, while the dilated convolution in MDFE focuses more on capturing local information and can better extract complex texture information in shallow-level features. The process of MDFE is as follows:

\begin{equation}
	\begin{aligned}
		F_1 &= CBL\left(F_{out}, rate = 1\right) \\
		F_3 &= CBL\left(F_{out}, rate = 3\right) \\
		F_6 &= CBL\left(F_{out}, rate = 6\right) \\
		F_{12} &= CBL\left(F_{out},rate = 12\right) \\
		S_{outj} &= CBL(CAT(F_1, F_3, F_6, F_{12}))
	\end{aligned}
	\label{mdfe-stepwise}
\end{equation}

\subsubsection{Deep-level Feature Iteration Module}

\begin{figure*}[htbp]
	
	\centerline{\includegraphics[width=45pc]{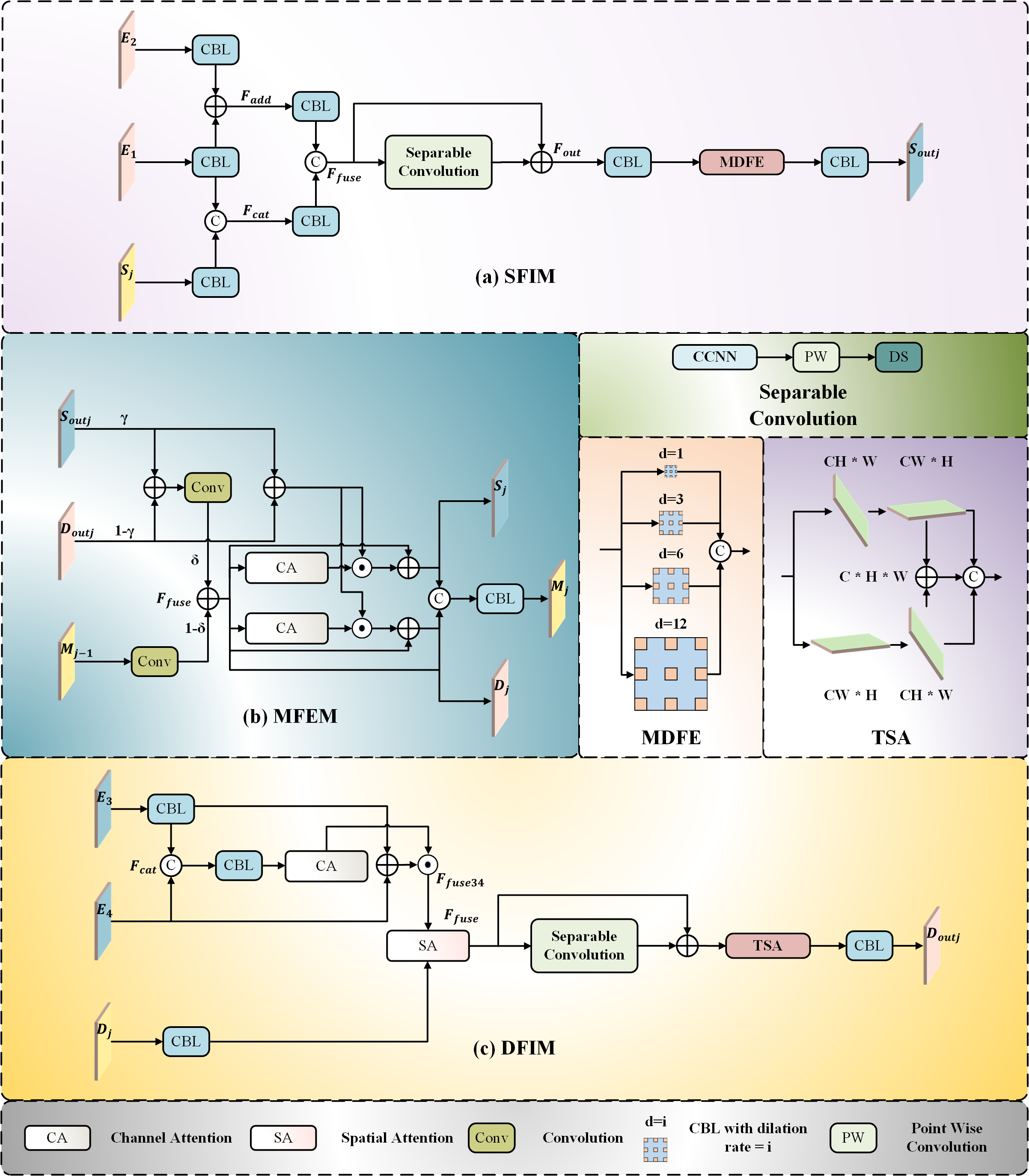}}
	\caption{Modules overall architecture: (a) The shallow-level feature iteration module (SFI-Module) is mainly used to extract and enhance the texture information in the shallow-level features. (b) The multi-feature enhancement module (MFE-Module) achieves the dynamic integration of shallow-level and deep-level features. (c) The deep-level feature iteration module (DFI-Module) processes deep-level features and extracts global semantic information.
	}
	\label{Module}
\end{figure*}

The DFI-Module is used to process deep-level features, and its architecture is shown in Fig. \ref{Module} (c). The input of this module is similar to SFI-Module, consisting of $E_3, E_4$, and $D_j$ (the first layer uses $E_4$ instead of $D_j$), where $E_3$ and $E_4$ represent deep-level features. The module upsamples the three inputs to adjust the size and then concatenates the two deep-level features. The concatenation procedure preserves the integrity of deep features and enhances their semantic representation. The features are then used in CA to enhance the representation of global semantics and reduce redundant information. Since deep-level features often include global semantic information, the supervision strategy used by the DFI-Module prioritizes the extraction of global semantics. Subsequently, we multiply the result of CA by the sum of $E_3$ and $E_4$, therefore enhancing the semantics of key channel features and enhancing nonlinear expression capabilities. Ultimately, we incorporate the result and $D_j$ into the SA to achieve the complementarity of semantic and spatial information.

\begin{equation}
	\begin{aligned}
		F_{cat} &= CAT(CBL(E_3), E_4) \\
		F_{fuse34} &= CA(CBL(F_{cat})) * (CBL(E_3) + E_4) \\
		F_{fuse} &= SA(F_{fuse34}, D_j)
	\end{aligned}
	\label{5}
\end{equation}

Subsequently, we introduced the separable convolution module and connected it with a two-dimensional splicing attention (TSA) to strengthen the interaction between channels and space. This module reintegrates the channel dimension and the spatial dimension through two-dimensional splicing and reconstruction. The process of the TSA module is as follows:

\begin{equation}
	\begin{aligned}
		X &= \text{Conv}(F_{out}) \\
		X_{1r} &= \text{Linear}_r(\text{Norm}_r(X^{(B \times CH \times W)}))\\
		X_{1c} &= \text{Linear}_c(\text{Norm}_c(X_{1r}^{(B \times WC \times H)}))\\
		X_{2c} &= \text{Linear}_c(\text{Norm}_c(X^{(B \times CW \times H)}))\\
		X_{2r} &= \text{Linear}_r(\text{Norm}_r(X_{2c}^{(B \times CH \times W)}))\\
		X_{12} &= X_{1c}^{(B \times C \times H \times W)} +X_{2r}^{(B \times C \times H \times W)} \\
		D_{outj} &= CAT(X_{12}, X_{1c}^{(B \times C \times H \times W)}, X_{2r}^{(B \times C \times H \times W)})
	\end{aligned}
\end{equation}

where Conv denotes convolution, Linear denotes linear transformation, and Norm represents normalization. The superscript of \( X \) (e.g., \( X^{(B \times CH \times W)} \)) represents the shape of the tensor at a specific stage, where \( B \) indicates the batch size, \( C \) indicates the number of channels, and \( H \) and \( W \) represent the height and width of the feature map, respectively. The TSA separates the input features into two branches: one that first expands the row dimension followed by the column dimension, and the other branch that expands the features in reverse order. The two results are then summed and concatenated with the initial input. Employing the TSA module to integrate deep-level features can effectively enhance the combination of global information, hence facilitating a more comprehensive understanding of the target's shape, location, and semantics.

\subsubsection{Multi Feature Enhancement Module}

Inspired by the cascaded multi-modal fusion unit, we propose a novel MFE-Module \cite{10064348}. This module receives three inputs: the outputs of SFI-Module and DFI-Module, $S_{outj}$ and $D_{outj}$, and the output of the previous MFE-Module, $M_{j-1}$ (the input of the first layer is replaced by $E_4$). Its architecture is shown in Fig. \ref{Module} (b).

Initially, we use the first learnable parameter to perform a weighted summation of the processed shallow-level and deep-level features, subsequently using the second parameter weight to sum up the result with $M_{j-1}$. The utilization of learnable parameters for double weighted summation enables dynamic adjustment of the fusion ratio, allowing for more flexible control over the contribution of the input of various levels of information. The process is as follows:

\begin{equation}
	\begin{aligned}
		F_{fuse} &= (1 - \delta) \cdot Conv(M_{j-1}) \\
		&\quad + \delta \cdot Conv(\gamma \cdot S_{outj} + (1 - \gamma) \cdot D_{outj})
	\end{aligned}
\end{equation}

The combined output is transmitted into the CA, and the result is multiplied by $(S_{outj} + D_{outj})$ to enhance the selectivity of significant semantic channels. Then the result is added to the original fused feature $F_{fuse}$ through the residual connection mechanism to ensure the integrity of the information and the robustness of the feature expression. This module employs two identical branch designs with CA, delivering the processed results to the SFI-Module and DFI-Module, respectively. This design effectively emphasizes the significant semantic channels, enabling the features to adapt to different feature representations in the subsequent fusion process. Finally, the results of the two branches are concatenated with $F_{fuse}$, and the features are integrated through the convolution layer to generate the enhanced semantic segmentation results. In the final semantic supervision part, the semantic segmentation results generated by the three MFE-Modules are added layer by layer for joint supervision. The process of generating the final three outputs is as follows:

\begin{equation}
	\begin{aligned}
		S_j &= CA(F_{fuse}) \cdot (S_{outj} + D_{outj}) + F_{fuse} \\
		D_j &= CA(F_{fuse}) \cdot (S_{outj} + D_{outj}) + F_{fuse} \\
		M_j &= CBL(CAT(F_{fuse}, S_j, D_j))
	\end{aligned}
\end{equation}

\subsection{Multi-Module Joint Supervision Strategy}

We use a multi-module joint supervision strategy, as shown in Fig. \ref{BIMII-Net}. We perform boundary supervision and binary supervision on the contour features output by the SFI-Module and the skeleton features output by the DFI-Module, respectively. Furthermore, we perform semantic supervision on the summation result of the three MFE-Modules. The supervision strategy we adopted regards multiple supervision heads as multi-task learning and uses learnable weights to perpetually optimize the weight values during training. We use the multi-task loss function proposed by Ref. \cite{Kendall_2018_CVPR} with cross-entropy loss for supervision:

\begin{equation}
	\begin{aligned}
		\mathcal{L}_{final} &= \sum_{i=1}^3 \frac{1}{2\sigma_i^2} \mathcal{L}_{bini} + \sum_{j=4}^6 \frac{1}{2\sigma_j^2} \mathcal{L}_{bou(j-3)} \\
		&\quad + \frac{1}{2\sigma_7} \mathcal{L}_{se} + \sum_{k=1}^7 \log \sigma_k
	\end{aligned}
\end{equation}

where $\sigma_k$$(k \in \{1, 2, 3, 4, 5, 6, 7\})$ represents the learnable weight, $\mathcal{L}_{bini}$ denotes binary supervision, $\mathcal{L}_{bou(j-3)}$ represents boundary supervision, and $\mathcal{L}_{se}$ denotes semantic supervision.
\section{Experiments}\label{sec4}

\subsection{Datasets and Evaluation Metrics}

We use two popular public datasets to evaluate BIMII-Net. The MFNet dataset is proposed by Ha et al. It records nine categories, including background, car, person, bike, curve, car stop, guardrail, color cone, and bump \cite{8206396}. The dataset contains 1569 pairs of RGB-T images acquired with an InfRec R500 camera, including 820 pairs of daylight scenes and 749 pairs of nocturnal scenes. The dataset has a resolution of $480 \times 640$, with a distribution ratio of 2:1:1 for the training set, validation set, and test set, respectively.

The second dataset is PST900, proposed by Shivakumar et al.\cite{shivakumar2020pst900}. This dataset has five categories: background, hand drill, backpack, fire extinguisher, and survivor, with a resolution of $720 \times 1280$. The dataset comprises 894 pairs of RGB-T pictures, partitioned into a training set of 597 pairs and a test set of 288 pairs.

We use two indicators to evaluate the performance of semantic segmentation: mean accuracy (mAcc) and mean intersection over union (mIoU). mAcc represents the average of the ratio of the number of correct pixels predicted by the model for each class to the number of all predicted pixels for that class. mIoU represents the average of the ratio of the intersection and union of the model's predicted results for each class and the true value. The formulas for these two evaluation indicators are as follows:

\begin{equation}
	\begin{aligned}
		\text{mAcc} &= \frac{1}{N} \sum_{i=1}^{N} \frac{\text{TP}_i}{\text{TP}_i + \text{FN}_i} \\
		\text{mIoU} &= \frac{1}{N} \sum_{i=1}^{N} \frac{\text{TP}_i}{\text{TP}_i + \text{FP}_i + \text{FN}_i}
	\end{aligned}
\end{equation}

where $N$ represents the number of classes, $\text{TP}_i, \text{FP}_i$ and $\text{FN}_i$ represent the true positives, false positives and false negatives in each class $i$, respectively.

\subsection{Implementation Details}

During training, we use random cropping and random flipping with a probability of 50\% for data augmentation. For the encoder part, we use the SegFormer-B3 model pre-trained on the ImageNet dataset \cite{xie2021segformer, 5206848}. We use Ranger as the optimizer and set a constant learning rate. We divide the training process into two stages: (a) The first stage is the direct training stage. In the first stage, the time step of each continuous-coupled neural network (CCNN) layer is 1 $(T=1)$.  We set the batch size to 2, the learning rate to $10^{-4}$, and the weight decay to $5 \times 10^{-4}$. (b) The second stage is the fine-tuning stage. We set the time step of the CCNN layer to 4 $(T=4)$, the batch size to 1, the learning rate to $10^{-5}$, the weight decay is unchanged, and the gradient clipping with a maximum norm of 20 is used. For the MFNet dataset, we use 75 epochs for direct training and 10 epochs for fine-tuning. For PST900 dataset, we use 110 epochs for direct training and 15 epochs for fine-tuning. Our experiments are all conducted on the Pytorch 2.3.0 framework and use an RTX 4090 graphics card with CUDA 12.4 and 24GB RAM.

\subsection{Comparative experiment}

\subsubsection{MFNet}

For the MFNet dataset, we compare BIMII-Net with the following methods: APCNet(4c)\cite{8954288}, DFN(4c)\cite{yu2018learning}, FRRN(4c)\cite{8099836}, BiSeNet(4c)\cite{10.1007/978-3-030-01261-8_20}, MFNet\cite{8206396}, FuseNet\cite{10.1007/978-3-319-54181-5_14} , ABMDRNet\cite{9578077}, RTFNet\cite{8666745}, MMNet\cite{lan2022mmnet}, GCNet\cite{liu2022gcnet}, DooDLeNet\cite{9856981}, RFIENet\cite{LI2022112177}, MMDRNet\cite{LIANG20239}, ECGFNet\cite{10041960} and MMSMCNet\cite{10123009}. Tab. \ref{comp} shows the quantitative results. It can be seen that the proposed BIMII-Net outperforms most other methods. Our method achieved four first places and six second places in the above results. Our results perform well on objects with small pixel appearance ratios such as "car stop", "guardrail" and "color cone", and the prediction effects of each category are relatively balanced, which reflects the robustness of the proposed method in the problem of class inconsistency.

\begin{table*}[ht]
	\centering
	\caption{Comparison of results on the MFNet dataset. The first and second place results of our network in each column are highlighted in red and blue, respectively.}
	\label{comp}
	\begin{adjustbox}{max width=\textwidth}
		\begin{tabular}{lcccccccccccccccccc}
			\toprule
			\textbf{Method} & \multicolumn{2}{c}{\textbf{Car}} & \multicolumn{2}{c}{\textbf{Person}} & \multicolumn{2}{c}{\textbf{Bike}} & \multicolumn{2}{c}{\textbf{Curve}} & \multicolumn{2}{c}{\textbf{Car Stop}} & \multicolumn{2}{c}{\textbf{Guardrail}} & \multicolumn{2}{c}{\textbf{Color Cone}} & \multicolumn{2}{c}{\textbf{Bump}} & \textbf{mAcc} & \textbf{mIoU} \\ 
			\cmidrule{2-17}
			& Acc & IoU & Acc & IoU & Acc & IoU & Acc & IoU & Acc & IoU & Acc & IoU & Acc & IoU & Acc & IoU & & \\ 
			\midrule
			APCNet(4c)\cite{8954288} & 86.6 & 81.2 & 74.3 & 59.8 & 64.6 & 55.9 & 42.0 & 33.0 & 36.8 & 25.1 & 33.2 & 7.1 & 47.6 & 37.5 & 55.7 & 51.8 & 60.0 & 49.9 \\ 
			DFN(4c)\cite{yu2018learning} & 90.0 & 84.4 & 73.2 & 65.0 & 75.5 & 60.9 & 54.0 & 40.4 & 38.9 & 25.7 & 10.2 & 4.0 & 48.3 & 42.5 & 55.8 & 47.4 & 60.5 & 52.0 \\ 
			FRRN(4c)\cite{8099836} & 81.9 & 74.7 & 66.2 & 60.8 & 62.8 & 50.3 & 41.2 & 35.0 & 12.5 & 11.5 & 0.0 & 0.0 & 37.2 & 34.0 & 35.2 & 34.6 & 48.5 & 44.2 \\ 
			BiSeNet(4c)\cite{10.1007/978-3-030-01261-8_20} & 89.7 & 84.1 & 72.0 & 63.2 & 74.1 & 60.1 & 45.1 & 36.7 & 34.2 & 25.3 & 18.2 & 5.0 & 47.4 & 42.2 & 39.8 & 35.9 & 57.7 & 50.0 \\ 
			MFNet\cite{8206396} & 77.2 & 65.9 & 67.0 & 58.9 & 53.9 & 42.9 & 36.2 & 29.9 & 12.5 & 9.9 & 0.1 & 0.0 & 30.3 & 25.2 & 30.0 & 27.7 & 45.1 & 39.7 \\ 
			FuseNet\cite{10.1007/978-3-319-54181-5_14} & 81.0 & 75.6 & 75.2 & 66.3 & 64.5 & 51.9 & 51.0 & 37.8 & 17.4 & 15.0 & 0.0 & 0.0 & 31.1 & 21.4 & 51.9 & 45.0 & 52.4 & 45.6 \\ 
			ABMDRNet\cite{9578077} & 94.3 & 84.8 & 90.0 & 69.6 & 75.7 & 60.3 & 64.0 & 45.1 & 44.1 & 33.1 & 31.0 & 5.1 & 61.7 & 47.4 & 66.2 & 50.0 & 69.5 & 54.8 \\ 
			RTFNet\cite{8666745} & 93.0 & 87.4 & 79.3 & 70.3 & 76.8 & 62.7 & 60.7 & 45.3 & 38.5 & 29.8 & 0.0 & 0.0 & 45.5 & 29.1 & 74.7 & 55.7 & 63.1 & 53.2 \\ 
			MMNet\cite{lan2022mmnet} & - & - & 83.9 & - & 69.3 & 59.0 & - & 43.2 & - & 24.7 & - & 4.6 & - & 42.2 & - & 50.7 & 62.7 & 52.8 \\ 
			GCNet\cite{liu2022gcnet} & 94.2 & 86.0 & 89.6 & 72.0 & 77.5 & 60.0 & 68.9 & 42.2 & 38.3 & 30.7 & 45.8 & 6.2 & 59.6 & 49.5 & 82.1 & 52.6 & 72.7 & 55.3 \\ 
			DooDLeNet\cite{9856981} & 91.7 & 86.7 & 81.3 & 72.2 & 76.6 & 62.5 & 58.9 & 46.7 & 36.2 & 28.0 & 35.2 & 5.1 & 56.9 & 50.7 & 74.8 & 65.8 & 67.9 & 57.3 \\ 
			RFIENet\cite{LI2022112177} & 92.0 & 87.5 & 81.1 & 73.0 & 78.7 & 62.5 & 62.9 & 44.7 & 44.2 & 34.3 & 39.3 & 7.9 & 62.9 & 44.5 & 79.0 & 60.6 & 71.0 & 57.0 \\ 
			MMDRNet\cite{LIANG20239} & 93.4 & 85.7 & 89.3 & 70.3 & 74.7 & 61.5 & 65.7 & 46.9 & 42.7 & 32.7 & 53.9 & 7.7 & 59.9 & 48.2 & 73.0 & 53.4 & 72.4 & 56.0 \\ 
			ECGFNet\cite{10041960} & 89.4 & 83.5 & 85.2 & 72.1 & 72.9 & 61.6 & 62.8 & 40.5 & 44.8 & 30.8 & 45.2 & 11.1 & 57.2 & 49.7 & 65.1 & 50.9 & 69.1 & 55.3 \\
			MMSMCNet\cite{10123009} & 96.2 & 89.2 & 93.2 & 69.1 & 83.4 & 63.5 & 74.4 & 46.4 & 56.6 & 41.9 & 26.9 & 8.8 & 70.2 & 48.8 & 77.5 & 57.6 & 75.2 & 58.1 \\ 
			BIMII-Net(Ours) & 93.3 & \textbf{\textcolor{blue}{87.7}} & 82.6 & \textbf{\textcolor{blue}{72.2}} & 75.9 & \textbf{\textcolor{blue}{63.0}} & 55.5 & 42.8 & \textbf{\textcolor{blue}{47.7}} & \textbf{\textcolor{blue}{37.3}} & \textbf{\textcolor{red}{54.3}} & \textbf{\textcolor{red}{11.8}} & 57.0 & \textbf{\textcolor{red}{50.8}} & 68.9 & \textbf{\textcolor{blue}{62.3}} & 70.5 & \textbf{\textcolor{red}{58.4}} \\
			\bottomrule
		\end{tabular}
	\end{adjustbox}
\end{table*}

We test RGB-T images during the day and at night in Tab. \ref{daynight}. It can be seen that although our model does not have a clear advantage in the daytime or nighttime indicators, its overall mIoU indicator exceeds that of most models. These results further demonstrate that our model has more balanced prediction performance during the day and at night and has stronger robustness.
\begin{table}[h!]
	\centering
	\caption{Results for daytime and nighttime images. The first and second place results of our network in each column are highlighted in red and blue.}
	\label{daynight}
	\begin{tabular}{lcccccc}
		\toprule
		\textbf{Methods} & \multicolumn{2}{c}{\textbf{Daytime}} & \multicolumn{2}{c}{\textbf{Nighttime}} & \multicolumn{2}{c}{\textbf{Test}} \\
		\cmidrule(lr){2-3} \cmidrule(lr){4-5} \cmidrule(lr){6-7}
		& \textbf{mAcc} & \textbf{mIoU} & \textbf{mAcc} & \textbf{mIoU} & \textbf{mAcc} & \textbf{mIoU} \\
		\midrule
		APCNet(4c) & 55.4 & 42.4 & 54.7 & 46.4 & 60.0 & 49.9 \\
		DFN(4c) & 53.7 & 42.2 & 52.4 & 44.6 & 60.5 & 52.0 \\
		FRRN(4c) & 45.1 & 40.0 & 41.6 & 37.3 & 48.5 & 44.2 \\
		BiSeNet(4c) & 52.1 & 44.5 & 50.3 & 45.0 & 57.7 & 50.0 \\
		MFNet & 42.6 & 36.1 & 41.4 & 36.8 & 45.1 & 39.7 \\
		RTFNet & 60.0 & 45.8 & 60.7 & 54.8 & 63.1 & 53.2 \\
		FuseNet & 49.5 & 41.0 & 48.9 & 43.9 & 52.4 & 45.6 \\
		DooDLeNet & 58.8 & 50.1 & 64.1 & 55.7 & 67.9 & 57.3 \\
		MMSMCNet & 81.2 & 56.9 & 72.2 & 58.7 & 75.2 & 58.1 \\
		Ours & \textbf{\textcolor{blue}{62.3}} & \textbf{\textcolor{blue}{52.4}} & \textbf{\textcolor{blue}{66.3}} & \textbf{\textcolor{blue}{56.1}} & \textbf{\textcolor{blue}{70.5}} & \textbf{\textcolor{red}{58.4}} \\
		\bottomrule
	\end{tabular}
\end{table}

\begin{figure*}[htbp]
	\centerline{\includegraphics[width=35pc]{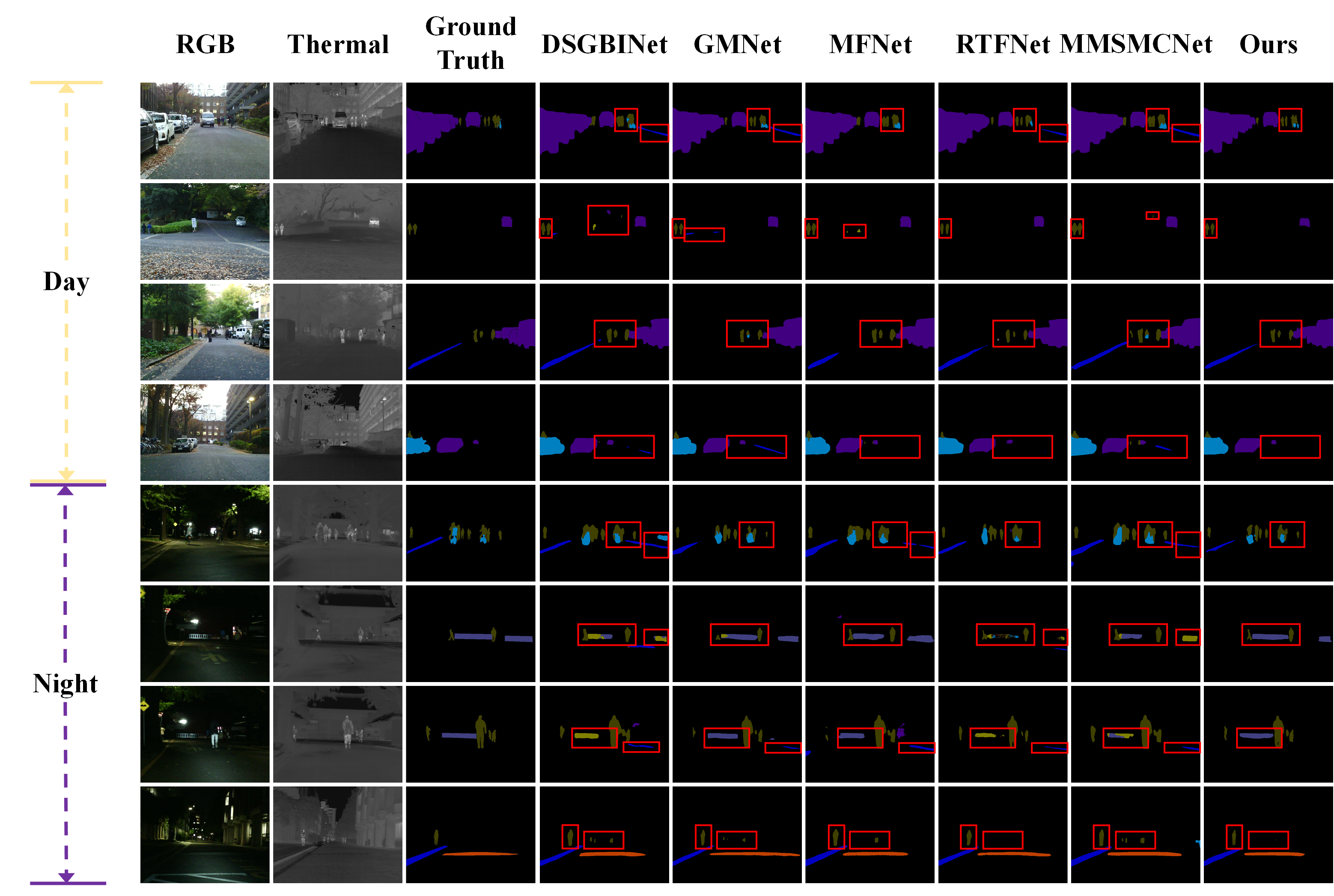}}
	\caption{Visualization results of multiple models on the MFNet dataset. The first four rows represent the segmentation results during the day, and the last four rows represent the segmentation results at night.}
	\label{com_mfnet}
\end{figure*}

The visualization results of the comparative experiment are shown in Fig. \ref{com_mfnet}, where the images used for comparison are divided into daytime and nighttime \cite{9531449, 9927165}. It can be seen that the predictions and actual results of BIMII-Net have high similarity and fewer misidentifications at night and during the day. Especially in the depiction of details, our model can more accurately identify local contour information such as the swinging arms of pedestrians, the distance between shoulders and legs, etc. In addition, our model has a high accuracy in predicting less frequent categories such as guardrails. Overall, our model has excellent prediction effects on both global skeleton information and local contour positioning.

\begin{figure}[htbp]
	\centerline{\includegraphics[width=22pc]{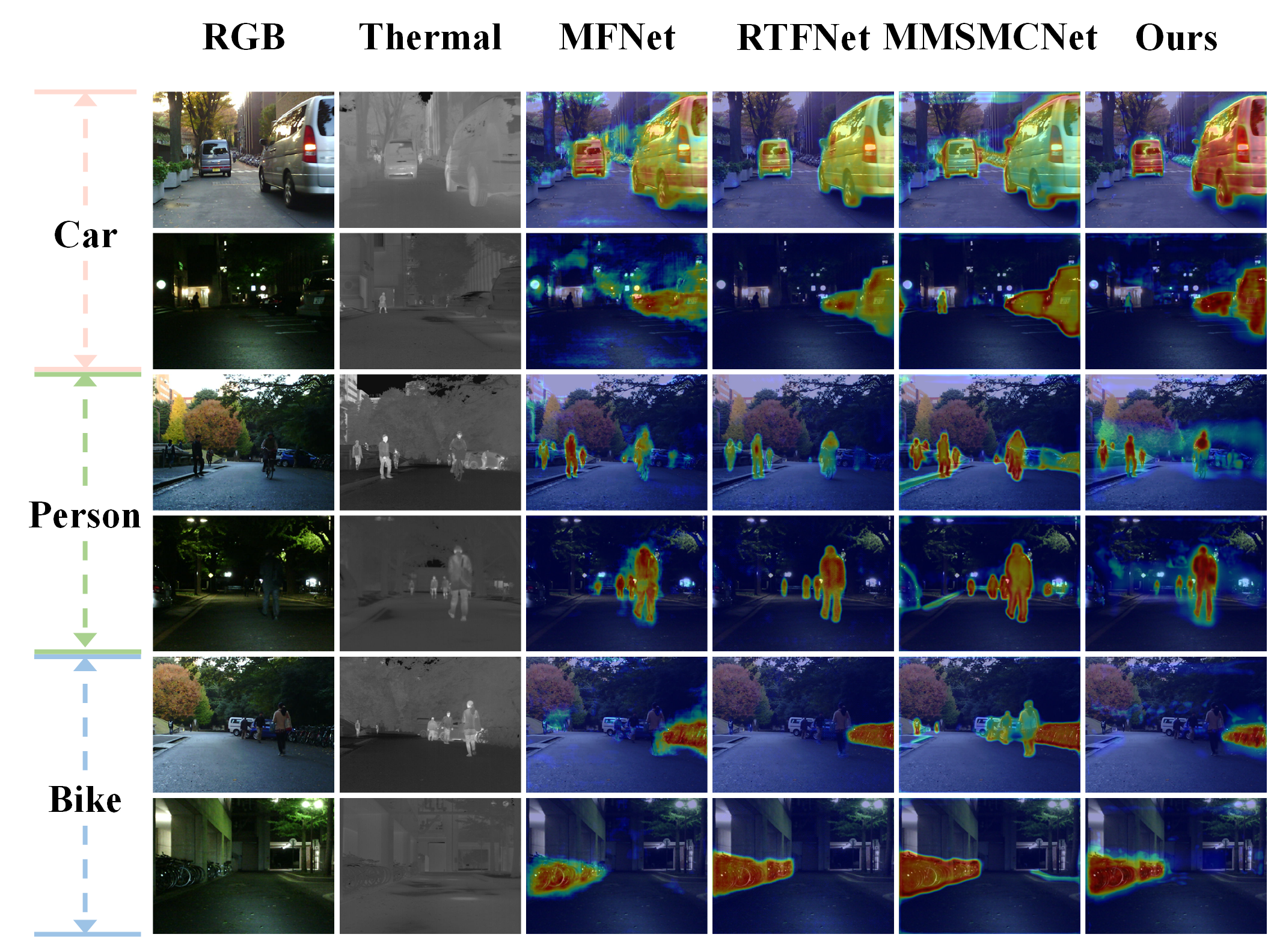}}
	\caption{Visualization of class activation maps of multiple models on the MFNet dataset, including three categories: car, person, and bike.}
	\label{cam}
\end{figure}

Fig. \ref{cam} shows the class activation map of our proposed model in three categories. It can be seen that the MMSMCMet pays more attention to the contour part of the category, and there is some feature confusion. MFNet and RTFNet are not clear enough for detecting boundaries. Overall, our model predicts more accurately and has clearer boundaries.

\subsubsection{PST900}

In the PST900 dataset, we compared our model with ERFNet\cite{8063438}, RTFNet\cite{8666745}, PSTNet\cite{shivakumar2020pst900}, CGFNet\cite{GUO2024111588}, MFFENet\cite{9447924}, MMDRNet\cite{LIANG20239}, FDCNet\cite{9987529}, and MMSMCNet\cite{10123009}. Tab. \ref{comp2} shows that our method achieved five first places and four second places in the above results, which is better than most other methods. Especially in the background and hand drill categories, our model achieved better results. Additionally, compared with other models, the indicators of each category of our model are more balanced.
\begin{table*}[ht]
	\centering
	\caption{Comparison of results on the PST900 dataset. The first and second place results of our network in each column are highlighted in red and blue, respectively.}
	\label{comp2}
	\begin{adjustbox}{max width=\textwidth}
		\begin{tabular}{lcccccccccccc}
			\toprule
			\textbf{Method} & \multicolumn{2}{c}{\textbf{Background}} & \multicolumn{2}{c}{\textbf{Hand-Drill}} & \multicolumn{2}{c}{\textbf{Backpack}} & \multicolumn{2}{c}{\textbf{Fire-Extinguisher}} & \multicolumn{2}{c}{\textbf{Survivor}} & \textbf{mAcc} & \textbf{mIoU} \\ 
			\cmidrule{2-11}
			& Acc & IoU & Acc & IoU & Acc & IoU & Acc & IoU & Acc & IoU & & \\ 
			\midrule
			ERFNet\cite{8063438}  & -     & 98.73 & - & 52.76 & - & 68.08 & -  & 58.79 & - & 34.38 & -   & 62.55 \\
			RTFNet\cite{8666745}  & 99.78 & 99.02 & 7.79  & 7.07  & 79.96 & 74.17 & 62.39 & 51.93 & 78.51 & 70.11 & 65.69 & 60.46 \\
			PSTNet\cite{shivakumar2020pst900}  & -     & 98.85 & -&53.60 & - & 69.20 & - & 70.12 & -     & 50.03 &-& 68.36 \\
			CGFNet\cite{GUO2024111588}  & 99.69 & 99.30 & 70.50 & 59.72 & 87.55 &82.00& 96.30 & 71.71 & 86.29 & 77.42 & 88.06&78.03 \\
			MFFENet\cite{9447924} & -     & 99.40 & -     & 72.50 & -     & 81.02     & - &66.38 & -     & 75.60 &-& 78.98 \\
			MMDRNet\cite{LIANG20239} & 99.50 & 98.90 & 77.10 & 40.60&79.10 & 71.10 & 77.90 & 52.40 & 69.80 & 62.30 &81.30& 68.70 \\
			FDCNet\cite{9987529}  & 99.72 & 99.15 & 82.52 & 70.36 & 77.45 & 72.17 & 91.77 & 71.52 & 78.36 & 72.36&85.96&77.11 \\
			MMSMCNet\cite{10123009}    & 99.55 &99.39 & 97.96 & 62.36&96.94&89.22&97.36&73.29&84.28&74.70&95.20&79.80 \\
			BIMII-Net(Ours) & \textbf{\textcolor{red}{99.78}} & \textbf{\textcolor{red}{99.41}} & \textbf{\textcolor{blue}{92.62}}& \textbf{\textcolor{red}{73.51}}&\textbf{\textcolor{blue}{90.56}} & \textbf{\textcolor{blue}{86.89}}& 92.86 & \textbf{\textcolor{red}{81.20}} & 75.92&72.26 & \textbf{\textcolor{blue}{90.37}} & \textbf{\textcolor{red}{82.65}} \\
			\bottomrule
		\end{tabular}
	\end{adjustbox}
\end{table*}

\subsubsection{Complexity}

We analyzed the complexity of our model in Tab. \ref{compl}, using four indicators: number of parameters, number of floating-point operations, model inference speed using GPU, and the amount of memory occupied by model parameters. It can be seen that compared with the models in recent years, the number of parameters of our model is lower than that of most models, but the number of floating-point operations and inference speed are higher than other models. Overall, the complexity of our model is still a certain distance away from true brain-like simulation. Our future work will continue to optimize the structure of the model and realize a model with low computational complexity.

\begin{table}[ht]
	\centering
	\caption{Comparison of complexity in multiple models.}
	\label{compl}
	\begin{tabular}{@{}lcccc@{}}
		\toprule
		\textbf{Models} & \textbf{Params/M} & \textbf{FLOPs/G} & \textbf{FPS} & \textbf{Size/MB} \\ 
		\midrule                 
		RTFNet & 254.51 & 337.46 & 24.08 & 970.88 \\
		MFNet & 0.74 & 8.42 & 193.96 & 2.81 \\
		FEANet & 255.21 & 337.47 & 27.24 & 973.54 \\
		GMNet & 153.01 & 149.90 & 40.79 & 583.70 \\
		MMSMCNet & 95.60 & 189.15 & 14.69 & 364.69 \\
		Ours & 99.46 & 260.40 & 12.80 & 379.41 \\
		\bottomrule
	\end{tabular}
\end{table}

\subsection{Ablation Experiment}

\subsubsection{Effectiveness of Each Module}

The upper part of Tab. \ref{abl} shows the ablation results of each module in BIMII-Net, where (w/o CEAEF) represents the ablation results of the cross explicit attention-enhanced fusion module (CEAEF-Module). The ablation experiments remove the CEAEF-Module, shallow-level feature iteration module (SFI-Module), deep-level feature iteration module (DFI-Module), multi-feature enhancement module (MFE-Module), and CCNN layer in turn. It can be seen that the three modules of the decoder and the CEAEF-Module all play an important role in our model. Without the MFE-Module, mAcc and mIoU are only 62.3 and 54.7, respectively, and the prediction effect decreases the most. After removing the CEAEF-Module, mAcc and mIoU are reduced to 65.1 and 55.0, respectively, indicating that this module can effectively enhance the fusion of global information. In addition, the ablation experiment on the CCNN layer reflects the role of the brain-inspired model in improving the model prediction effect.

\begin{table}[ht]
	\centering
	\caption{Results of ablation experiments. The upper part of the table shows the ablation results of different modules, the middle part shows the ablation results of different structures within the modules, and the lower part shows the ablation results of different supervision strategies.
	}
	\label{abl}
	\begin{adjustbox}{}
		\begin{tabular}{lcc}
			\toprule
			\textbf{Models} & \textbf{mAcc} & \textbf{mIoU} \\
			\midrule
			Model (w/o CEAEF) & 65.1 & 55.0 \\
			Model (w/o SFI-Module) & 66.8 & 54.3 \\
			Model (w/o DFI-Module) & 67.1 & 55.3 \\
			Model (w/o MFE-Module) & 62.3 & 54.7 \\
			Model (w/o CCNN) & 68.6 & 56.1 \\
			\midrule
			Model (w EAEF) & 68.2 & 58.1 \\
			Model (SFI-Module w/o MDFE) & 61.9 & 53.8 \\
			Model (SFI-Module w/o CCNN) & 63.4 & 53.5 \\
			Model (DFI-Module w/o TSA) & 68.1 & 56.0 \\
			Model (DFI-Module w/o SA) & 64.0 & 55.3 \\
			Model (DFI-Module w/o CCNN) & 64.1 & 56.3 \\
			\midrule
			Model (w/o $loss_{bin}$) & 66.7 & 54.8 \\
			Model (w/o $loss_{bou}$) & 62.4 & 54.9 \\
			Model (w/o $loss_{1, 2}$) & 68.4 & 57.3 \\
			Model (w/o $loss_{1, 3}$) & 66.9 & 55.6 \\
			Model (w/o $loss_{2, 3}$) & 66.4 & 55.3 \\
			Model (w/o $loss_{1, 2, 3}$) & 65.0 & 55.2 \\
			Model (w loss weight(3,1,1)) &65.2& 54.8 \\
			Model (w loss weight(1,1,1)) &65.1& 53.6 \\
			\textbf{Model (Ours)} & \textbf{70.5} & \textbf{58.4} \\
			\bottomrule
		\end{tabular}
	\end{adjustbox}
\end{table}

\begin{figure*}[htbp]
	\centerline{\includegraphics[width=40pc]{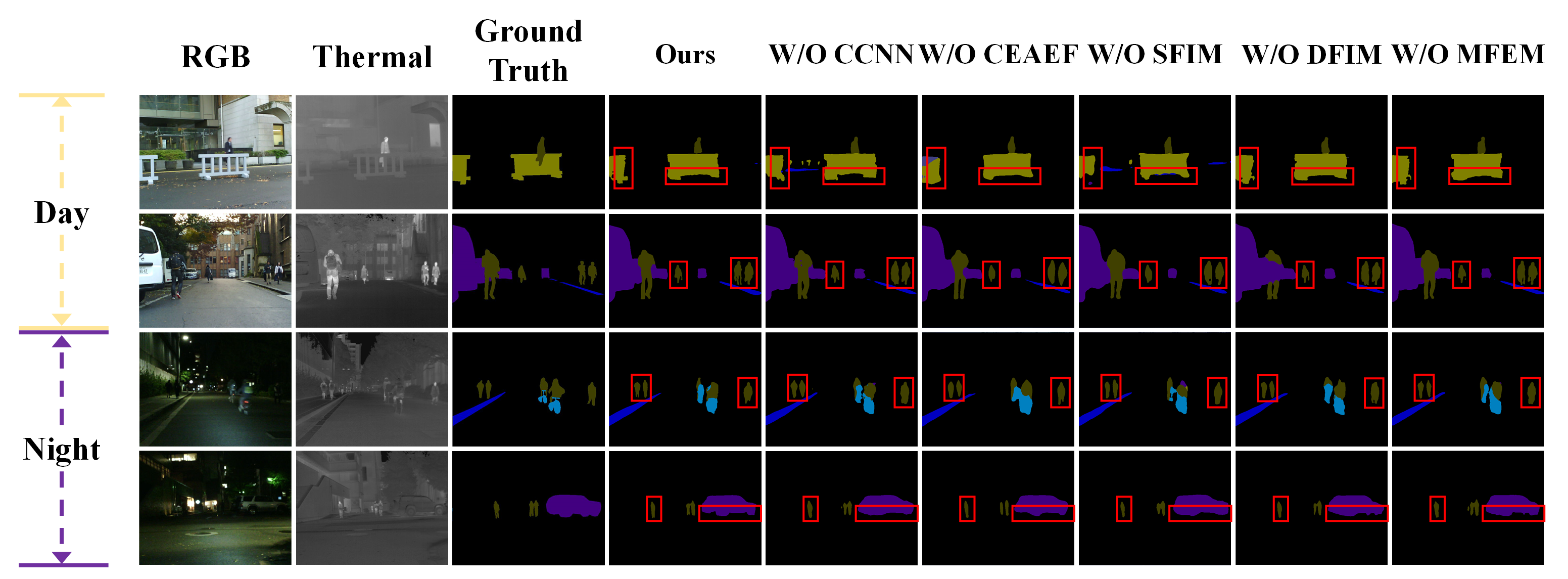}}
	\caption{Visualization of ablation experiment results for different modules.
	}
	\label{abl2}
\end{figure*}

We also visualized the ablation results of each key module, as shown in Fig. \ref{abl2}. It can be seen that when some modules are missing, the model's contour is not clear and misrecognition occurs. These problems are particularly prominent when the decoder module is missing. Our model better depicts details such as the car's tires and swinging arms, reflecting the importance of each module.

\subsubsection{Validity of the Internal structure of Modules}

To further verify the rationality of module design, we conducted a detailed ablation experiment on the internal structure of modules, and the results are shown in the middle part of Tab. \ref{abl}. The experimental design includes replacing or removing certain submodules: (a) (w EAEF) means replacing the CEAEF-Module with the EAEF module, (b) (SFI-Module w/o MDFE) means removing the multi-scale dilated feature extraction (MDFE) operation in the SFI-Module, (c) (SFI-Module w/o CCNN) means removing the CCNN layer in the SFI-Module, (d) (DFI-Module w/o TSA) means removing the two-dimensional splicing attention (TSA) in the DFI-Module, (e) (DFI-Module w/o SA) means removing the SA in the DFI-Module, and (f) (DFI-Module w/o CCNN) means removing the CCNN layer in the DFI-Module. The experiments show that the improved CEAEF-Module outperforms the traditional EAEF module in global information fusion and effectively improves the prediction effect. In addition, MDFE in the SFI-Module is the key to performance improvement. After removing it, the model performance drops most significantly (61.9/53.8). Besides, the use of CCNN layers in SFI-Module and DFI-Module also effectively improves the model's prediction ability. These experimental results verify the positive role of our proposed new module and improved design in improving model performance.

\subsubsection{Effectiveness of Supervision Strategy}

The lower part of Tab. \ref{abl} confirms the effectiveness of the multi-module joint supervision strategy we adopted. The experimental design includes removing the binary loss $loss_{bin}$, the boundary loss $loss_{bou}$, the loss supervision of each layer $loss_{1, 2, 3}$, and using fixed loss weights. The experiment shows that $loss_{bin}$ and $loss_{bou}$ play a key role in global localization and boundary optimization, respectively. After removing them, their mAcc and mIoU are reduced to 66.7/54.8 and 62.4/54.9, respectively. Additionally, the loss supervision of each layer ($loss_{1, 2, 3}$) can provide effective supervision on features at different levels. After all are removed, the model performance drops significantly (mAcc and mIoU drop to 65.0/55.2). In addition, the use of fixed loss weights also has a significant impact on performance, where the weight (3, 1, 1) means:

\begin{equation}
	\sigma_k = 
	\begin{cases} 
		1, & k \in \{1, 2, 3, 4, 5, 6\} \\
		3, & k = 7
	\end{cases}
\end{equation}

where $\sigma_k$ represents a fixed weight, which achieves 65.23 and 54.75 on mAcc and mIoU, respectively, which is better than the fixed weight (1, 1, 1), but still lower than the multi-task loss function we used. These experimental results demonstrate the effectiveness and rationality of our proposed multi-module joint supervision strategy.
\subsection{Effectiveness of CCNN}

At present, no model has been found that uses brain-inspired computing for multi-modal semantic segmentation. In the comparative experiment of spiking neural network (SNN) models, we tried to use the most advanced SNN architecture to conduct experiments, which is shown in Tab. \ref{snn}. The experimental results show that the existing brain-inspired computing architecture performs poorly in multi-modal semantic segmentation tasks, and its prediction effect is much lower than that of the proposed architecture based on CCNN. The BIMII-Net proposed by us is better than the current state-of-the-art method (SOTA) in brain-inspired models.

To further verify the effectiveness of CCNN, we used different brain-inspired models (pulse-coupled neural network (PCNN) and LIF) to replace CCNN for experiments \cite{nelson1995hodgkin}. Among them, Nolinking CCNN and Nolinking PCNN represent CCNN and PCNN with $ \beta = 0$ and $V_F = 0$, respectively. Their characteristics are that the internal neurons are independent of each other and do not perform coupling operations. Compared with traditional CCNN, Nolinking CCNN has lower computational complexity but cannot simulate the characteristics of biological neurons more accurately.

%


For the non-differentiable problem of spike output, we use the alternative gradient method proposed by SpikingJelly\cite{doi:10.1126/sciadv.adi1480}. The results show that the Nolinking state of PCNN has better performance, with mAcc and mIoU of 65.5 and 56.5, respectively, which is the best performance of the brain-inspired model using spike output in the experiment. In addition, the indicators of the traditional LIF model are the worst compared with other models, with mAcc and mIoU of only 41.7 and 32.8. It can be found that the prediction effect of the brain-inspired models with spike output in the table (PCNN, Nolinking PCNN and LIF) in the state of multiple iterations $(T=4)$ is slightly reduced, indicating that these models are not suitable for too many iterations. In general, the prediction effects of the brain-inspired models listed in Tab. \ref{snn} are lower than those of CCNN, indicating that CCNN has better performance advantages when processing multi-modal data.

\begin{table}[ht]
	\centering
	\caption{Comparison of results of different brain-inspired computing models on the MFNet dataset.}
	\label{snn}
	\begin{adjustbox}{}
		\begin{tabular}{lccc}
			\toprule
			\textbf{Models} &T& \textbf{mAcc} & \textbf{mIoU} \\
			\midrule
			Meta-SpikeFormer \cite{yao2024spikedriven} &4& 12.9 & 11.0 \\
			Spiking-Unet \cite{LI2024127653} &-& 11.1 & 10.3 \\
			\multirow{2}{*}{Model (w Nolinking PCNN)} &1& 65.5 & 56.5 \\
			&4& 64.9 & 56.3 \\
			\multirow{2}{*}{Model (w PCNN)} &1& 68.0 & 56.5 \\
			&4& 64.7 & 55.7 \\
			\multirow{2}{*}{Model (w Nolinking CCNN)} &1& 70.2 & 57.0 \\
			&4& 69.4 & 57.5 \\
			\multirow{2}{*}{Model (w LIF)} &1& 41.7 & 32.8 \\
			&4& 40.1 & 31.7 \\
			\multirow{2}{*}{Model (w CCNN)} &1& 67.2 & 57.6 \\
			&4& \textbf{70.5} & \textbf{58.4} \\
			\bottomrule
		\end{tabular}
	\end{adjustbox}
\end{table}

We conducted an ablation experiment on the size of the convolutional layer used in CCNN in Tab. \ref{conv}. It can be seen that the convolutional layer size of $7 \times 7$ shows the best performance, followed by the dilated convolution of $5 \times 5$, whose mAcc and mIoU are 68.7 and 56.3, respectively. The experimental results show that the size of the convolutional layer has a significant impact on the feature modeling ability and final performance of the model.

\begin{table}[ht]
	\centering
	\caption{Comparison of results on the MFNet dataset using convolution kernels of different sizes in CCNN.}
	\label{conv}
	\begin{adjustbox}{}
		\begin{tabular}{lccc}
			\toprule
			\textbf{Models} &Conv Size& \textbf{mAcc} & \textbf{mIoU} \\
			\midrule
			\multirow{4}{*}{Model (w CCNN)} &$3 \times 3$& 66.2 & 56.0 \\
			&$5 \times 5$, 2 dilated& 68.7 & 56.3 \\
			&$5 \times 5$& 64.8 & 55.5 \\
			&$7 \times 7$, 3 dilated& 66.0 & 55.2 \\
			&$7 \times 7$& \textbf{70.5} & \textbf{58.4} \\
			\bottomrule
		\end{tabular}
	\end{adjustbox}
\end{table}

\subsection{Failure Case}

\begin{figure}[htbp]
	\centerline{\includegraphics[width=22pc]{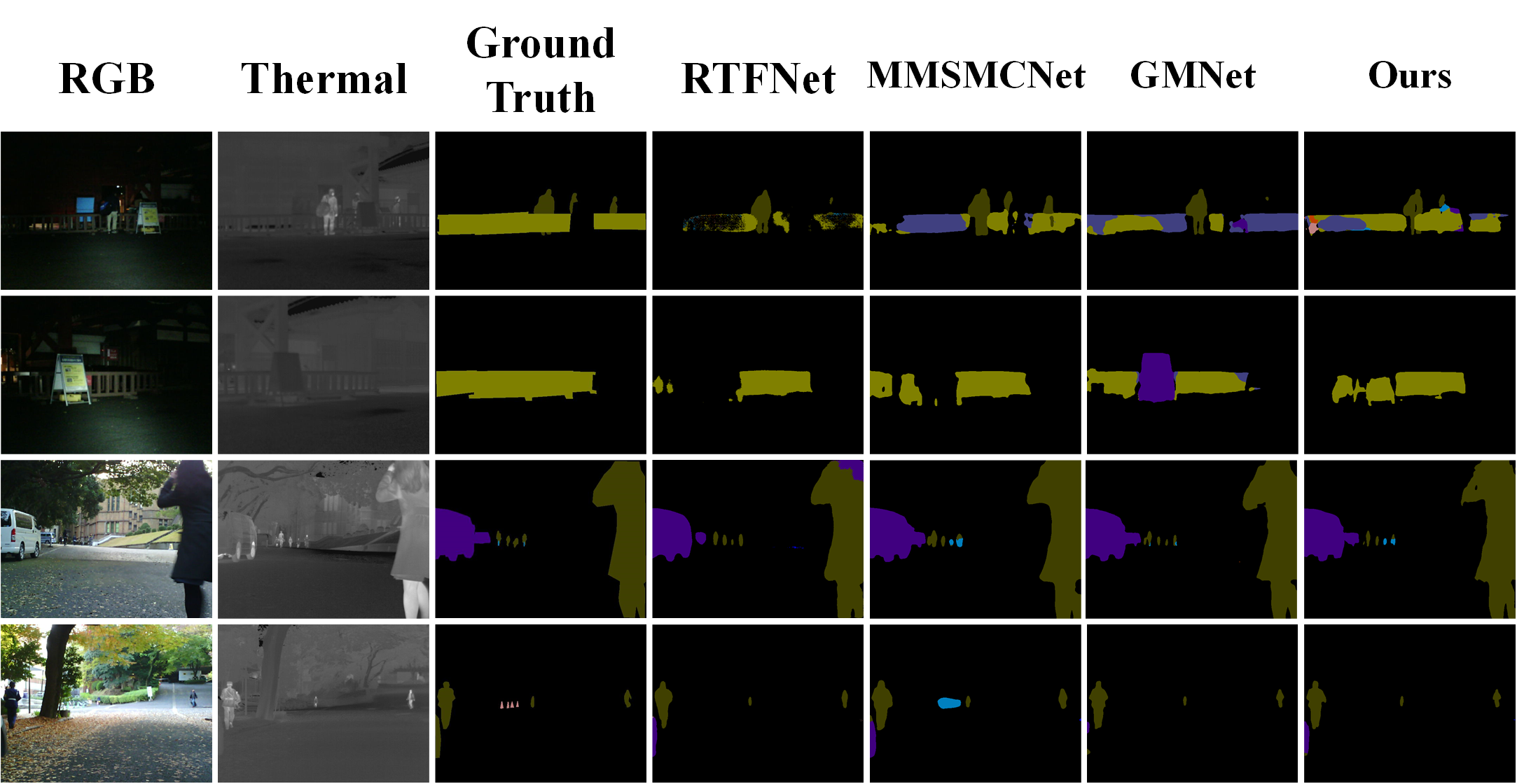}}
	\caption{Four sets of failure cases for multiple models.}
	\label{fail}
\end{figure}

We found that there are several groups of failed cases when using the proposed BIMII-Net, as shown in Fig. \ref{fail}. It can be seen that the first row of Fig. \ref{fail} misidentifies the stop sign as a guardrail or other category, and there is a case of misrecognition, while the second row fails to fully recognize the stop sign, and there is a case of insufficient recognition. The misrecognition of these two rows is caused by the occlusion of the target. In addition, the third and fourth rows have misrecognition of pedestrians and bicycles and omission of color cones, respectively, which are all caused by the small size of the target. However, compared with other models, it can be seen that these problems also exist in them, and most models are more serious than our proposed model.

\section{Conclusion}\label{sec5}

In this paper, we propose a novel RGB-T road scene semantic segmentation model, Brain-Inspired Multi-Iteration Interaction Network (BIMII-Net). The proposed deep continuous-coupled neural network (DCCNN) utilizes the iterative optimization mechanism of the brain-inspired model continuous-coupled neural network (CCNN) to facilitate layer-by-layer fusion and progressive updates of multi-scale features, thus effectively capturing texture and local detail information. In addition, the improved cross explicit attention-enhanced fusion module (CEAEF-Module) we developed fully integrates multi-modal information from RGB and thermal images through a complementary branch design, significantly enhancing the interaction between different features. Furthermore, the complementary interactive multi-layer decoder utilizes the integration of the shallow-level feature iteration module (SFI-Module), the deep-level feature iteration module (DFI-Module), and the multi-feature enhancement module (MFE-Module) to facilitate the complementary extraction of global skeleton and local detail information, thereby enhancing the segmentation efficacy. Experiments across multiple datasets confirm the superior performance of the proposed method, and comprehensive ablation experiments demonstrate the effectiveness of the architecture and modules. However, the complexity and computational requirements of the model may still impose specific constraints in practical applications. In future work, we aim to optimize the architecture of BIMII-Net to develop a lightweight and energy-efficient model suitable for real-world deployment.

\bibliography{refs}
\textbf{\bibliographystyle{IEEEtran}}

\begin{IEEEbiography}[{\includegraphics[width=1in,height=1.25in,clip,keepaspectratio]{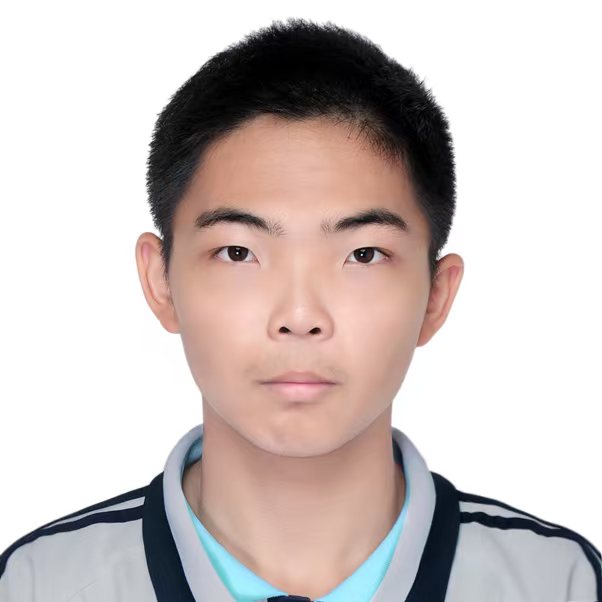}}]{Hanshuo Qiu}
	born 2003 in Fuzhou, Fujian, China, is currently pursuing a B.S. degree at the School of Information Science and Engineering, Lanzhou University. His research mainly focuses on chaotic systems, image encryption, and deep learning.
\end{IEEEbiography}

\begin{IEEEbiography}[{\includegraphics[width=1in,height=1.25in,clip,keepaspectratio]{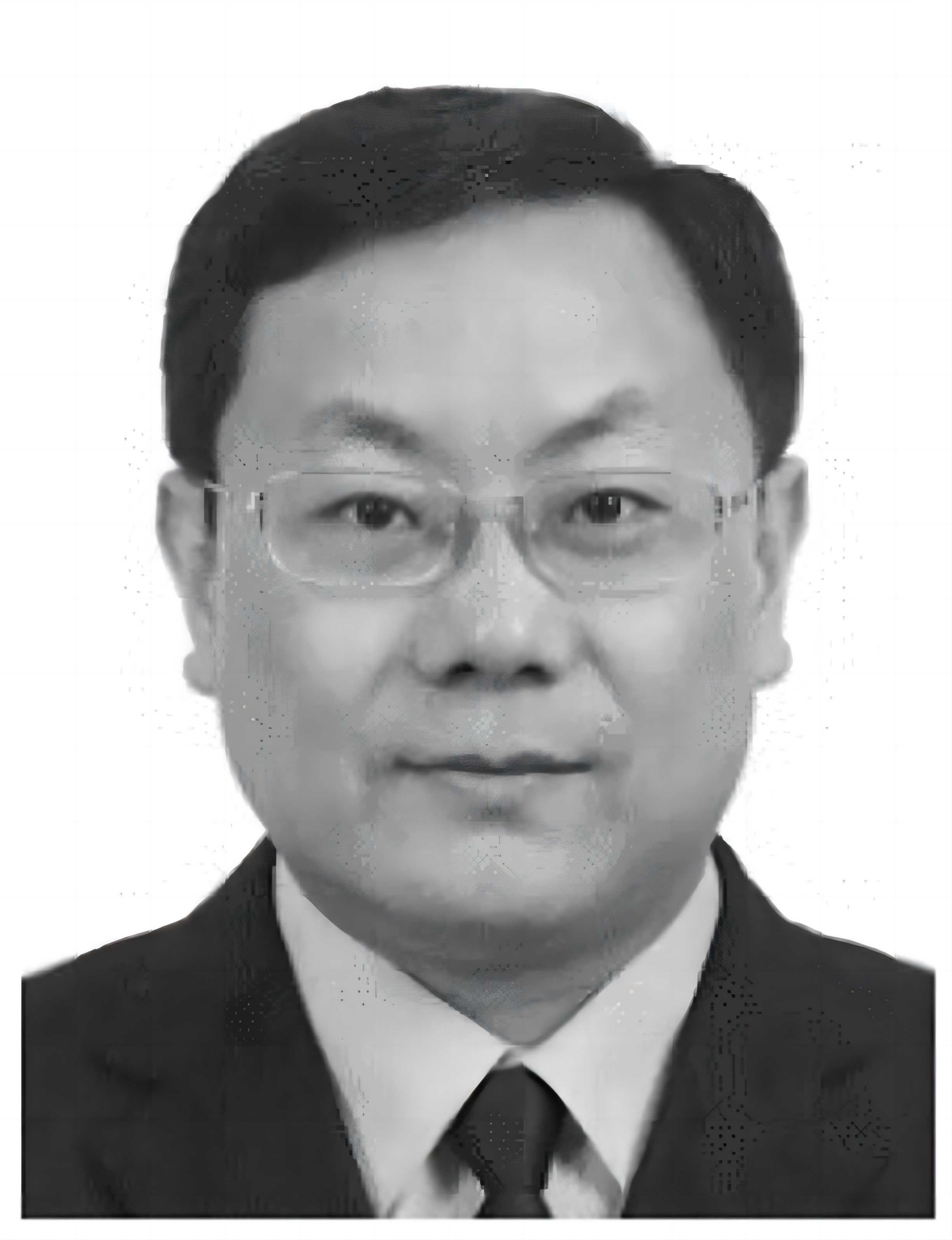}}]{Jie Jiang}
	is a professor at the Key Laboratory of Big Data and Decision Making, Department of Systems Engineering, National University of Defense Technology. His research interests include artificial intelligence and deep learning, visualization and visual analytics, virtual reality and intelligent interaction.
\end{IEEEbiography}

\begin{IEEEbiography}[{\includegraphics[width=1in,height=1.25in,clip,keepaspectratio]{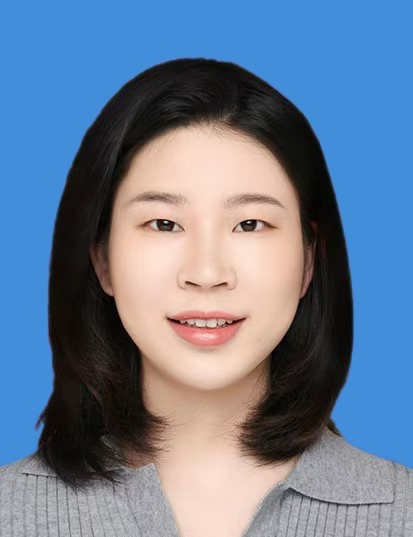}}]{Ruoli Yang}
	is a passionate and dedicated individual currently pursuing her education at the National University of Defense Technology. With a strong inclination towards technology and a thirst for knowledge, she has chosen to focus her academic pursuits on the fields of deep learning, graphics, and image processing. 
	She is driven by a desire to contribute to the development of advanced algorithms and techniques that can enhance the capabilities of artificial intelligence systems.
\end{IEEEbiography}

\begin{IEEEbiography}[{\includegraphics[width=1in,height=1.25in,clip,keepaspectratio]{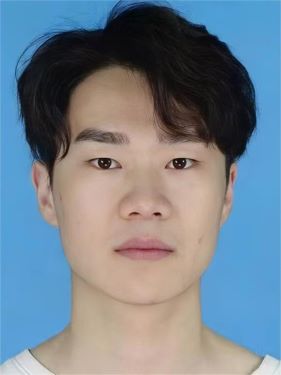}}]{Lixin Zhan}
	obtained the M.S. degree in computer technology from Nanchang University, China, in 2024. He is currently pursuing a Ph.D. degree at National University of Defense Technology, China. His research interests include 3D computer vision and deep learning.
\end{IEEEbiography}

\begin{IEEEbiography}[{\includegraphics[width=1in,height=1.25in,clip,keepaspectratio]{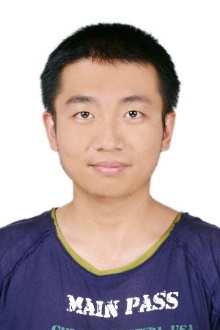}}]{Jizhao Liu}
	born 1989 in Lanzhou, Gansu, China, received his B.S, M.S. and Ph.D. in Lanzhou University. He was a postdoctoral fellow at Sun Yat-sen University. He is an associate professor with School of Information Science and Engineering, Lanzhou University, China. His research has been primarily in the area of chaos theory, chaotic circuit, and related engineering applications. His current research interests include the nonlinear characteristic in biological neural network, neural coding, brain-like and brain-inspired computing.
\end{IEEEbiography}
\end{document}